\documentclass{article}

\usepackage{PRIMEarxiv}
\usepackage[utf8]{inputenc} 
\usepackage{graphicx}
\usepackage{subcaption}
\usepackage[colorlinks=true,linkcolor=blue,citecolor=blue]{hyperref}
\usepackage{xr}  

\usepackage{comment}
\usepackage{soul}
\usepackage{xcolor}

\usepackage{booktabs}
\usepackage{natbib}%
 \usepackage{graphics}
\usepackage{tabularx}
\usepackage{makecell}
\usepackage[export]{adjustbox}
\usepackage{multirow}
\usepackage{enumitem}
\setlist{labelindent=12pt}
\usepackage{subcaption}
\usepackage{algorithm} 
\usepackage{algpseudocode}
\usepackage{float}
\usepackage[labelfont=bf,textfont={bf}]{caption}
\usepackage{xcolor}
\usepackage{nccmath, mathtools}
\usepackage{svg}
\usepackage{amssymb} 
\usepackage{etoolbox}
\usepackage{threeparttable}
\apptocmd\normalsize{%
 \abovedisplayskip=12pt
 \abovedisplayshortskip=0pt 
 \belowdisplayskip=12pt 
 \belowdisplayshortskip=7pt
}{}{}
\usepackage{mathtools}
\usepackage{physics}
\hypersetup{
    colorlinks=true,                
    breaklinks=true,                
    urlcolor= cyan,                
    linkcolor= red,                     
    bookmarksopen=false,
    filecolor=black,
    citecolor=blue,
    linkbordercolor=pink
}

\renewcommand{\mathbf}{\boldsymbol}

\newcount\Comments  
\definecolor{darkgreen}{rgb}{0,0.5,0}
\definecolor{purple}{rgb}{1,0,1}
\definecolor{darkblue}{rgb}{0.6,0.4,0.8}

\newcommand{\kibitz}[2]{\ifnum\Comments=0\textcolor{#1}{#2}\fi}

\usepackage{caption}
\title{Large Language Models for Pedestrian Safety:  An Application to Predicting Driver Yielding Behavior at Unsignalized Intersections

\thanks{\textit{\underline{Citation}}: 
\textbf{Yang, et al. Large Language Models for Pedestrian Safety:  An Application to Predicting Driver Yielding Behavior at Unsignalized Intersections.}} 
}

\author{
  Yicheng Yang \\
  School of Artificial Intelligence\\
  Hebei University of Technology  \\
   \And
  Zixian Li \\
  School of Artificial Intelligence\\
  Hebei University of Technology  \\
 \AND
Jean Paul Bizimana \\
  Department of Civil Engineering \\
  Saint Louis University \\
    \AND
    Niaz Zafri \\
  Department of Urban Studies and Planning \\
  Massachusetts Institute of Technology\\
   \AND
    Yongfeng Dong \\
  Hebei Province Key Laboratory of Big Data Computing\\
  Hebei University of Technology  \\
  \AND
  Tianyi Li \\
  Department of Civil Engineering \\
  Saint Louis University \\
  \texttt{tianyili.ai@gmail.com} \\}

\begin{document}
\maketitle

\begin{abstract}

Pedestrian safety is a critical component of urban mobility and is strongly influenced by the interactions between pedestrian decision-making and driver yielding behavior at crosswalks. Modeling driver–pedestrian interactions at intersections requires accurately capturing the complexity of these behaviors. 
Traditional machine learning models often struggle to capture the nuanced and context-dependent reasoning required for these multifactorial interactions, due to their reliance on fixed feature representations and limited interpretability. In contrast, large language models (LLMs) are suited for extracting patterns from heterogeneous traffic data, enabling accurate modeling of driver-pedestrian interactions.
Therefore, this paper leverages multimodal LLMs through a novel prompt design that incorporates domain-specific knowledge, structured reasoning, and few-shot prompting, enabling interpretable and context-aware inference of driver yielding behavior, as an example application of modeling pedestrian–driver interaction.
We benchmarked state-of-the-art LLMs against traditional classifiers, finding that GPT-4o consistently achieves the highest accuracy and recall, while Deepseek-V3 excels in precision.
These findings highlight the critical trade-offs between model performance and computational efficiency, offering practical guidance for deploying LLMs in real-world pedestrian safety systems.

\end{abstract}


\section{Introduction}\label{sec1}

Pedestrian safety has become an increasingly urgent concern in urban traffic systems. 
In the United States, pedestrian fatalities increased by 12.5 percent from 2020 to 2021 and reached 7,388 deaths, which represents the highest level recorded in several decades~\citep{NHTSA}.
This alarming trend underscores the need to reassess existing safety measures, particularly in settings where pedestrian–vehicle interactions are frequent and risks are elevated~\citep{craig2019pedestrian}.

One of the most critical environments for such interactions is the intersection. Pedestrian safety in these locations depends on the behavior of two primary actors: pedestrians and drivers. On the pedestrian side, crossing behavior reflects how individuals weigh risk, value their time, interpret traffic cues, and adopt various strategies to navigate the roadway~\citep{kadali2012pedestrians,zafri2023walk}. 
Risky crossing choices, such as misjudging traffic gaps, substantially increase the likelihood of crashes~\citep{piyalungka2025pedestrian}.

On the driver's side, yielding behavior plays an equally decisive role. This challenge is particularly acute at unsignalized intersections, where the absence of traffic control devices places responsibility directly on both road users~\citep{LIU2025271,BONERA2024}. The willingness of drivers to respond directly shapes the outcomes of the interaction. Failure to yield not only forces pedestrians to pause or retreat but also greatly increases the risk of pedestrian–vehicle conflicts, often resulting in severe injury or death~\citep{Golembiewski2011}. Thus, accurately understanding and predicting the behaviors of both pedestrians and drivers is vital for designing targeted safety interventions and for supporting early warning systems, intelligent intersections, and future autonomous vehicle technologies.

Recent advances in LLM open new possibilities for addressing these challenges. LLMs are a subclass of generative AI based on transformer architectures, and have rapidly expanded beyond natural language processing to domains such as traffic safety~\citep{zhang2024advancing}. They capture complex contextual relationships and integrate heterogeneous inputs, including textual reports, sensor data streams, and visual cues, to generate accurate predictions~\citep{chang2024survey,hurst2024gpt}. These capabilities position them as powerful tools for modeling pedestrian–vehicle interactions. Specifically, multimodal LLMs can interpret visual cues such as pedestrian body language, vehicle positioning, and traffic conditions, alongside textual and sensor data, to infer likely interaction outcomes, including driver yielding decisions and pedestrian crossing behaviors. This capacity makes LLMs especially promising for pedestrian safety applications in real-time contexts, where subtle dynamics often determine crash risk.

Although LLMs hold broad potential for improving pedestrian safety, this study focuses on a specific application that demonstrates this potential by predicting driver yielding behavior toward pedestrians at unsignalized intersections.
Existing research has disproportionately emphasized pedestrian risk taking behaviors (e.g., jaywalking, distraction, and decision-making)~\citep{Ghanbari2024,jayatilleke2022analysing,zafri2023walk}, while comparatively little attention has been paid to driver yielding behavior and its dependence on broader environmental and contextual factors. 
To date, most studies have relied primarily on statistical classification models (e.g., logistic regression) to investigate how individual factors, including road geometry, traffic volume, vehicle speed, roadside land use, and visibility, influence driver yielding decisions~\citep{Schroeder2011,li2025identifying,zafri2022effect}.
However, the interplay among these factors remains insufficiently explored. By leveraging LLMs, we assess their predictive performance across diverse contexts, explore their ability to capture complex interdependencies, and establish a foundation for future applications of LLMs in pedestrian safety research and practice.

This study advances the literature by evaluating the capacity of LLMs to predict driver yielding behavior at unsignalized intersections, drawing on more than 3,000 annotated pedestrian–driver interaction instances across 18 unsignalized intersections in Minnesota. Our contributions are threefold:

\begin{itemize}
  \item Novel application of multimodal LLMs: we design prompts that integrate domain-specific knowledge, structured reasoning guidance, and few-shot examples, enabling interpretable and context-aware inference of driver yielding behavior.
  
  \item Comprehensive model evaluation: we compare multiple LLMs from two major model families (the GPT series and the DeepSeek framework), alongside traditional classification methods (support vector machine (SVM), random forest (RF), neural network (NN) and logistic regression (LR)), analyzing predictive accuracy, interpretability, and computational efficiency in heterogeneous traffic contexts.

  \item Actionable insights for deployment: through fine-grained comparative analysis, we provide practical guidance and identify future research directions for deploying LLMs in pedestrian safety systems, contributing to the development of intelligent, context-aware traffic management technologies.
  
\end{itemize}

The rest of this paper is organized as follows. Section \ref{Literature Review} reviews literature related to AI and LLM in traffic safety and driver yielding behavior. Section \ref{Data} presents the characteristics and descriptive statistics of the adopted dataset. Section \ref{Results from Traditional Classifier} describes the traditional classifier selected after comparison. Section \ref{Methodology} introduces the proposed predictive framework along with its core components and methodology. Section \ref{Experiment} evaluates the performance of various predictive models, followed by a comparative analysis in Section \ref{Discussion}. Section \ref{Future} discusses the study’s limitations, highlights key challenges, and identifies directions for future research. Finally, Section \ref{Conclusion} concludes the paper.

\section{Literature review}\label{Literature Review}

Recent advances in AI have created significant opportunities to enhance traffic safety through data-driven methodologies. In particular, generative AI and LLMs have demonstrated considerable potential in reasoning across heterogeneous traffic data. Meanwhile, driver yielding behavior at unsignalized intersections continues to be a critical area of traffic safety research, given its direct impact on pedestrian risk and overall traffic efficiency. This section reviews previous studies on the application of generative AI and LLMs in traffic safety, and subsequently summarizes empirical research on driver yielding behavior and the key factors influencing driver decision-making.

\subsection{Generative AI and LLMs in traffic safety}

Recent advances in generative AI and LLMs have opened promising avenues for improving traffic safety within intelligent transportation systems (ITS). By enabling more intelligent and context-aware decision-making, these models can support applications such as optimized signal timing, intersection and vehicle warning systems, driver assistance technologies, data collection, and enhanced traffic operations. When integrated into existing AI-driven frameworks, LLMs hold the potential to advance safer and more efficient urban mobility.
Although direct applications of LLMs in traffic safety remain rare and at an early stage, traditional machine learning (ML) and deep learning (DL) approaches have long been adopted to support traffic management and safety analysis. Techniques such as RF, bioinspired algorithms, and autonomic computing, time series forecasting, multilayer perceptrons, SVM, and LR have been applied to detect accidents and generate alternative routes~\citep{thomas2017toward,frank2019multilayer,lozano2020analysis}, analyze crash severity~\citep{bhuiyan2022crash,ahmadi2020crash}, and analyze driver and pedestrian behavior~\citep{li2025identifying,qian2010support,zhu2023investigation}. These models have also shown effectiveness in predicting pedestrian crossing intentions~\citep{singh2024prediction} and driver yielding behavior~\citep{li2025identifying,negash2023driver}. 
Despite such progress, persistent challenges, including model interpretability, rigid input requirements, limited ability to capture complex interdependencies, and bias in training datasets, continue to constrain the robustness and reliability of these models~\citep{hahn2019security}.
The use of LLMs in traffic safety research, though limited, has recently gained attention. 
Recent work has developed a framework that leverages LLM capabilities to analyze traffic crash narratives and identify underreported contributing factors~\citep{arteaga2025large}. This framework integrates multiple components, including prompt definition, selection of LLM generation parameters, output parsing, and underreporting determination. To evaluate its effectiveness, the case of identifying underreported alcohol involvement in traffic crashes was examined. The study further investigates the detection accuracy of the framework across different underlying LLMs (including ChatGPT, Flan-UL2, and Llama-2), prompt designs (explicit vs. implicit matching), and text generation parameters (e.g., sampling temperature and nucleus sampling probability). It demonstrates the potential of LLMs to identify latent patterns in narrative data that traditional statistical or rule-based approaches may overlook. 
In another study that utilized LLM capabilities to analyze traffic crash narratives, ChatGPT, BARD, and GPT-4 were employed to automatically extract information from 100 crash reports from Iowa and Kansas~\citep{mumtarin2023large}. The research evaluated the performance of these models across five key questions: fault identification, manner of collision, workzone involvement, pedestrian involvement, and sequence of harmful events. The results indicated high inter-model consistency on binary classification tasks such as workzone (96\%) and pedestrian involvement (89\%), whereas significantly lower agreement (35\%) was observed for more complex tasks such as collision manner. For event sequence analysis, network modeling revealed structural discrepancies among the outputs of different models, though some shared central events were identified. This study recommends using multiple LLM in combination and advises caution when extracting safety-critical information from narrative data.
Beyond such analysis, the generative capabilities of LLM have been leveraged to simulate realistic traffic scenarios, enabling virtual testing of safety interventions prior to field implementation~\citep{singh2024prediction}.

LLMs are also becoming increasingly relevant to decision-making in automated vehicle (AV) systems. Transformer-based models integrated with reinforcement learning have been used to predict pedestrian trajectories and adjust AV yielding behavior accordingly~\citep{wang2023drivemlm}. This capability is especially important in dense urban environments, where unsignalized intersections introduce significant uncertainty. Continuous training on real-world scenarios has further improved model responsiveness, supporting safer interactions with pedestrians and optimizing yielding decisions in both simulated and experimental contexts~\citep{cui2024large}.

Nevertheless, several limitations persist. Data bias can reduce model performance in underrepresented or adverse conditions~\citep{tan2023leveraging,hahn2019security}, while reliance on sensitive real-time inputs such as video feeds and location data raises privacy concerns. Federated learning has emerged as a promising solution, retaining up to 89\% model accuracy while preserving 98\% of data locally, thus mitigating privacy risks in sensitive environments such as school zones and residential neighborhoods~\citep{hahn2019security}. Furthermore, integrating LLM with complementary technologies, such as computer vision, LiDAR, and edge computing, has demonstrated strong potential to enhance real-time decision-making, provide timely alerts to drivers, and thus improve pedestrian safety~\citep{yan2024survey,iyer2021enabled,lozano2020analysis}.

\subsection{Driver yielding behavior}

While extensive research identifies key factors in driver yielding, modeling their complex interplay remains a challenge for traditional methods. Modern LLMs are uniquely positioned to address this gap. Their ability to perform nuanced contextual reasoning offers a powerful new paradigm for creating more holistic and interpretable models of driver-pedestrian interactions. Driver yielding behavior at unsignalized intersections is a critical determinant of pedestrian safety and traffic flow efficiency. Understanding how and why drivers decide to yield provides valuable insights for designing effective interventions, ranging from infrastructure improvements to intelligent transportation systems. This section first introduces the concept and importance of yielding behavior in traffic safety research, followed by a review of the key factors that influence drivers’ decisions in such contexts.

\subsubsection{Understanding driver yielding behavior and its importance}

Driver yielding behavior is generally defined as the outcome of a pedestrian–vehicle interaction, which occurs when a pedestrian and an approaching vehicle meet in close temporal and spatial proximity at an intersection, requiring at least one party to adjust their movement trajectory or speed in response to the other~\citep{zafri2022effect}. These interactions are typically categorized into two types: yield events, in which the driver alters speed or stops to allow the pedestrian to cross, and non-yield events, in which the pedestrian either waits for the vehicle to pass or crosses despite the absence of driver compliance. 

Yielding behavior plays a critical role in safeguarding pedestrians, cyclists, and other vulnerable road users~\citep{bella2021drivers,wang2021investigating}, especially at unsignalized intersections or locations where traffic signals provide insufficient control. Understanding and analyzing this behavior is essential because it informs the design of safer road systems, supports more effective traffic management strategies, advances driver assistance technologies, and reduces the likelihood of pedestrian crashes. Accurately predicting driver yielding behavior is critical for optimizing real-time traffic control and supporting AV decision-making~\citep{dey2021communicating}. Understanding human driver responses helps AVs navigate unpredictable pedestrian movements safely, enhancing both road safety and the seamless integration of AVs into urban traffic systems~\citep{yang2019contributes,albdairi2024impact,lee2020gentlemen}.

\subsubsection{Factors influencing drivers’ decisions to yield at unsignalized intersections}

Accurately modeling pedestrian–vehicle interactions at unsignalized intersections requires a comprehensive understanding of the contextual factors shaping driver yielding behavior. Such behavior emerges from a complex interplay of vehicle dynamics, environmental conditions, and human characteristics. From the vehicle perspective, approach speed and traffic volume are critical determinants. Lower speed is consistently associated with higher yielding rates~\citep{mdlmnDOT2024,bertulis2014driver}, whereas higher speed and heavier traffic volumes often reduce compliance due to increased driver stress and congestion~\citep{gankhuyag2023naturalistic,ito2017predicting}.

Environmental and infrastructural features also exert a substantial influence. Drivers exhibit a higher likelihood of yielding in environments characterized by elevated pedestrian activity, particularly in proximity to schools, parks, and urban centers~\citep{li2025identifying}.
Design interventions, including elevated crosswalks, visible markings, and traffic calming measures, have been shown to increase yield rates~\citep{Guo2019,schneider2015pedestrian}, while wider roadways and adjacent parking facilities may discourage compliance~\citep{aziz2013exploring,zahabi2011estimating}. Adverse environmental conditions, such as poor street lighting, nighttime settings, fog, and rain, reduce pedestrian visibility and consequently lower the likelihood of drivers yielding~\citep{badri2016influence}.

In addition to situational and environmental factors, individual characteristics of drivers and pedestrians significantly shape outcomes. Older and more experienced drivers tend to be more cautious and demonstrate higher yielding rates, while younger or less experienced drivers often exhibit lower compliance~\citep{yang2020psychological}. Psychological factors, including attention, perception, and risk sensitivity, further influence driver decision-making in real time~\citep{yang2020psychological}. Pedestrian characteristics also affect yielding behavior: individuals who act cautiously or are highly conspicuous are more likely to prompt yielding responses from drivers~\citep{zafri2022effect, Tian2020}. Ultimately, the perceived vulnerability of pedestrians combined with drivers’ attentional states determines the dynamics of these interactions, highlighting the importance of incorporating behavioral insights into predictive models of yielding behavior.

\section{Experimental data and settings}\label{Data}

This study draws on the open source Minnesota driver–pedestrian dataset documented in \cite{Stern2023, li2025identifying, li2021leveraging}, comprising 3,314 interactions recorded at 18 unsignalised intersections. These sites were specifically chosen in collaboration with the Minnesota Department of Transportation to ensure a diverse representation of traffic volumes, land use contexts, and intersection geometries. The complete annotated dataset is publicly available for research purposes~\citep{stern2023naturalistic}.

\subsection{Data pre-processing methodology}
\begin{figure}[!htb]
    \centering
    \includegraphics[width=\linewidth]{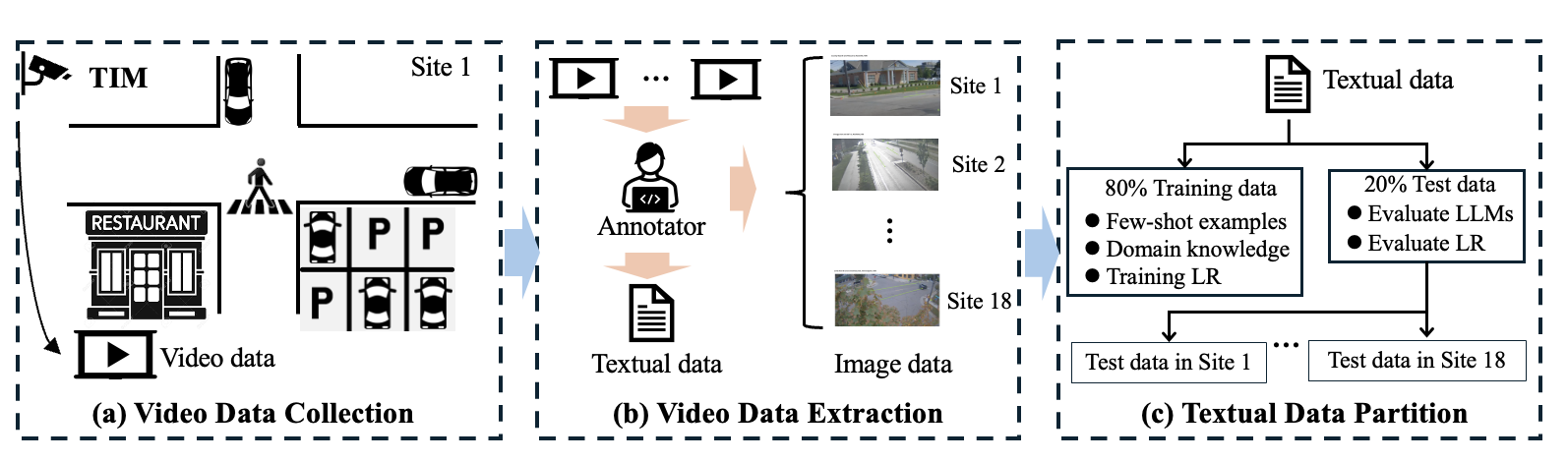}
    \caption{Procedures to pre-process driver–pedestrian interaction data.}
    \label{fig:Data processing}
\end{figure}

Data used for LLM requires specific data pre-processing procedures in Fig.~\ref{fig:Data processing}.

In Step (a), raw video data were collected during the summer and fall of 2021 using the Traffic Information Monitor (TIM) platform, a customized sensor suite developed by the Minnesota Traffic Observatory~\citep{li2025identifying}.
Each TIM unit consists of a pole-mounted video camera extendable up to 30 feet, a Raspberry Pi computer, and a battery pack allowing for extended operation. To capture naturalistic driver and pedestrian behaviors, the units were deployed for two weeks at each site to record video during daylight hours, with no supplemental signage or enforcement used. 

In Step (b), all video footage captured by the TIM units was manually reviewed by one trained annotator, who identified and labeled all relevant interaction events. These interactions are typically categorized into two types: yield events, in which the driver alters speed or stops to allow the pedestrian to cross, and non-yield events, in which the pedestrian either waits for the vehicle to pass or crosses despite the absence of driver compliance. This manual annotation process ensured consistency and accuracy throughout the dataset. Two forms of input data were extracted during annotation. The textual data consists of textual descriptions that captured the contextual and behavioral characteristics of each interaction. For each intersection, one of the most representative images was selected as an additional multimodal input. While the textual data already provide a comprehensive description of both the dynamic interaction processes and static contextual features of each event, the inclusion of imagery aims to enhance the LLM’s understanding of the intersection’s specific conditions. This visual information allows the model to capture subtle environmental and structural details that may be difficult to convey through text alone, thereby supporting more nuanced and informed reasoning. 
%
For instance, images can reveal conditions such as faded crosswalk markings, the presence of turn-only lanes, or potential obstacles like roadside trees.

In Step (c), the textual data were divided into training and testing sets. Noteworthy, LLMs do not require a traditional training process, and thus the training set was primarily used to construct few-shot exemplars and incorporate relevant domain knowledge. The test set was further partitioned by intersection site to assess the generalization of model predictions across diverse physical and behavioral contexts.

\begin{table}
    \aboverulesep=0pt
    \scriptsize
    \renewcommand{\arraystretch}{1.0}
    \centering
    \caption{Categorization of selected variables.}
    \resizebox{\textwidth}{!}{
    \begin{tabular}{ccc}
    \hline
   \multicolumn{2}{c}{\textbf{Category}} &  \textbf{Variables} \\
    \hline
    \multicolumn{2}{c}{
    \text{
    Vehicle dynamics and control
    }
    }&  \begin{tabular}{p{6cm}}
    Vehicle speed, Opposite direction yield (the yielding behavior of vehicles traveling in the opposite direction), Speed limit
    \end{tabular}
    \\
    \hline
    ~& 
    \begin{tabular}{p{2.5cm}}
    Road characteristics (geometric and design features of the road)
    \end{tabular}&\begin{tabular}{p{6cm}}
    Crossing width, Bike lanes, Signage, Markings,
 Number of bus stops
    \end{tabular}\\
    \cmidrule(r){2-3}
    \begin{tabular}{p{1.2cm}}
    Road networks and infrastructure
    \end{tabular}
    & 
    \begin{tabular}{p{2.5cm}}
    Presence of buildings (adjacent land use indicative of building presence) 
    \end{tabular}
    & \begin{tabular}{p{6cm}} 
    Presence of single family housing, Presence of apartments, Presence of commercial buildings, Presence of gas station/convenience store, Presence of restaurants/bars, Presence of parking lots, Presence of on street parking
    \end{tabular}\\
    \cmidrule(r){2-3}
    ~ & 
    \begin{tabular}{p{2.5cm}}Distance in miles from public facilities 
    \end{tabular}&\begin{tabular}{p{6cm}}
    Distance to nearest park, Distance to nearest school
    \end{tabular} \\ 
    \hline
     \multicolumn{2}{c}{
    \text{
    Pedestrian mobility and interaction
    }
    }& \begin{tabular}{p{6cm}}
    Number of pedestrians, Pedestrian type (an individual's mode of movement, e.g., walking, biking, using a scooter, or accompanying elements, e.g., with a dog, stroller, or child, in traffic environments)
    \end{tabular}
    \\
    \hline
    \end{tabular}
    }
    \label{Categorization of Variables Used in the Study}
\end{table}

\subsection{Dataset overview and site characteristics}

The carefully curated dataset initially comprised over 50 contextual variables. However, some variables were excluded due to either irrelevance to the research objectives or potential leakage of the yielding outcome~\citep{Dataset}. After this refinement, a total of 19 variables were retained for the subsequent analysis and classified into three primary categories in Table \ref{Categorization of Variables Used in the Study}. The category of road networks and infrastructure is further delineated into three subgroups to enhance the interpretability of variables.

The adopted dataset is collected from 18 selected sites that exhibit significant diversity in their design and operational characteristics. Table~\ref{Site Characteristics} summarizes these key site attributes, including roadway geometry, posted speed limits, annual average daily traffic (AADT), and the total number of recorded events at each site. This variability ensures that the dataset captures a wide range of real-world scenarios.

\begin{table}[h]
    \centering
    \renewcommand{\arraystretch}{1.0}
    \caption{Site characteristics.}
    \begin{tabular}{lcccccc}
\hline
\textbf{Site} & \textbf{Lanes} & \textbf{Speed limit} & \textbf{Markings} & \textbf{AADT} & \textbf{Shape} & \textbf{Event Count} \\
\hline
Site 1  & 4 & 35 & Unmarked    & 14,600 & T-shape & 31\\
Site 2  & 2 & 35 & Unmarked    & 14,800 & T-shape & 100\\
Site 3  & 2 & 30 & Standard   & 14,800 & T-shape & 640\\
Site 4  & 2 & 30 & Unmarked    & 10,700 & Four-way & 143\\
Site 5  & 2 & 30 & Unmarked    & 10,500 & Four-way & 341\\
Site 6  & 3 & 35 & Continental & 6,200  & Four-way & 95\\
Site 7  & 4 & 35 & Continental & 3,300  & T-shape & 121\\
Site 8  & 4 & 30 & Unmarked    & 10,200 & Four-way & 105\\
Site 9  & 3 & 35 & Unmarked    & 5,400  & Four-way & 74 \\
Site 10 & 4 & 35 & Unmarked    & 18,800 & Four-way & 88\\
Site 11 & 3 & 30 & Unmarked    & 18,100 & Four-way & 20\\
Site 12 & 2 & 30 & Unmarked    & 10,300 & Four-way & 122\\
Site 13 & 2 & 35 & Continental & 7,400  & Four-way & 98\\
Site 14 & 2 & 30 & Standard    & 7,100  & Four-way & 133\\
Site 15 & 3 & 30 & Unmarked    & 16,400 & Four-way & 46\\
Site 16 & 2 & 30 & Standard    & 12,100 & Four-way & 841\\
Site 17 & 4 & 30 & Unmarked    & 12,800 & T-shape & 127\\
Site 18 & 4 & 30 & Unmarked    & 12,500 & Four-way & 189\\
\hline
\end{tabular}
\label{Site Characteristics}
\end{table}

\begin{figure}[!hb]
    \centering
    \includegraphics[width=0.85\linewidth]{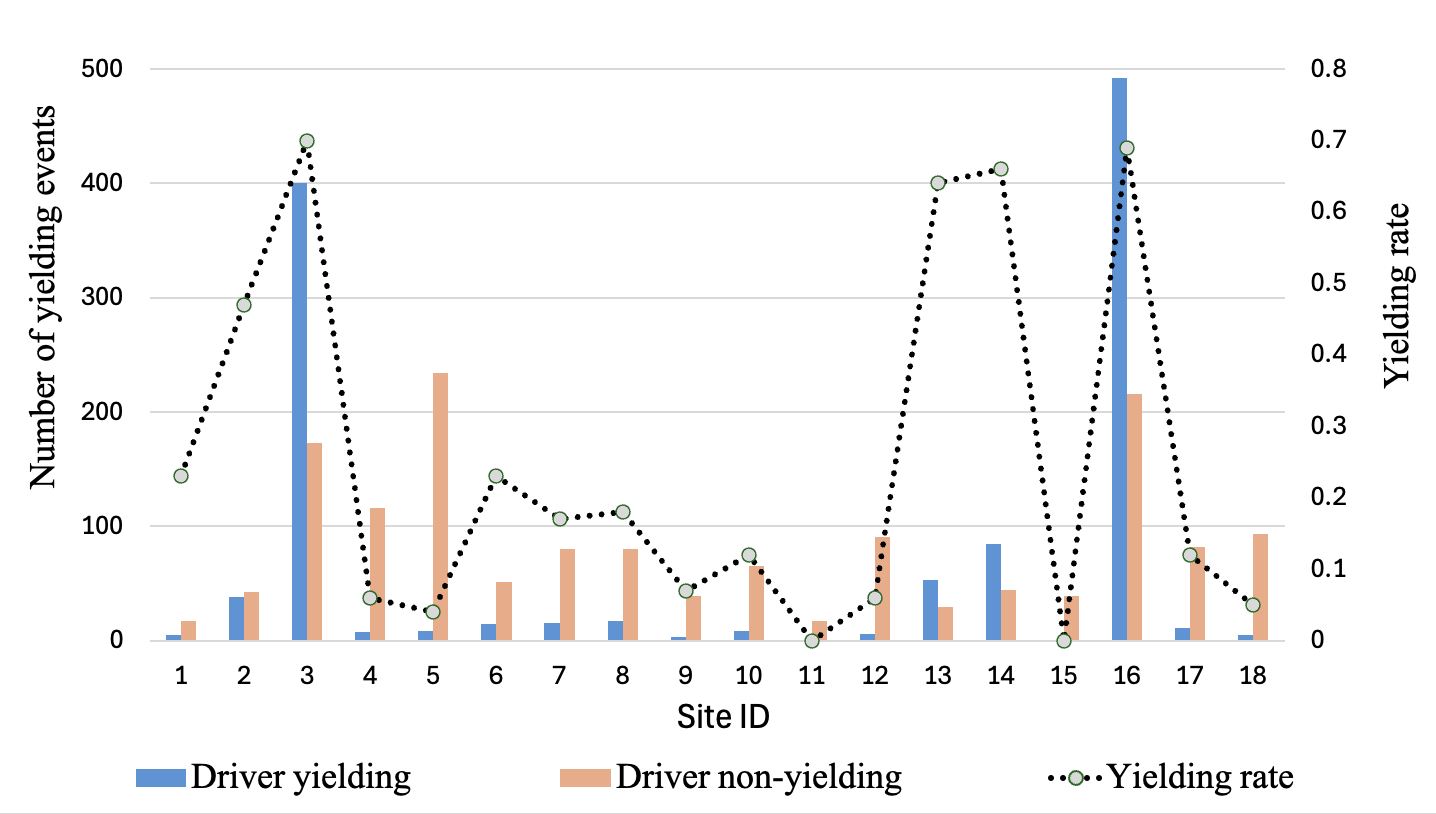}
    \caption{Distribution of driver yielding and non-yielding events across sites with corresponding overall yielding rates.}
    \label{fig:yielding_rates}
\end{figure}

Fig.~\ref{fig:yielding_rates} illustrates the number of driver-yielding and non-yielding events across the 18 studied intersections, along with their corresponding overall yielding rates. In general, yielding rates remain low at most study sites. 
Site 16 recorded the highest yielding rate of 70.36\% with a total of 841 events, whereas sites 11 and 15 reported no driver-yielding events among 20 and 46 events, respectively. The absence of yielding behavior at these sites may reflect local environmental conditions that lower drivers’ expectations or the necessity to yield to pedestrians. These findings suggest that intersections with higher pedestrian traffic volumes are more likely to promote yielding behavior among drivers.

\subsection{Descriptive statistics}\label{Descriptive Statistics}

This section performs preliminary analysis for variables in the adopted dataset. Table~\ref{Crosstabs of the summary categorical statistics between yield and non-yield events} and Table~\ref{Crosstabs of the summary numerical statistics between yield and non-yield events} respectively present the summary statistics of categorical variables and numerical variables.
These variables are classified into three categories. In the category of vehicle dynamics and control, three variables are of importance in yielding and non-yielding events:
(1) \textbf{vehicle speed}: during yielding events, vehicle speed averaged 10.0 MPH, with the 25th and 75th percentiles at 5.6 MPH and 12.6 MPH, respectively. In contrast, vehicle speed during non-yielding events was markedly higher, with a mean of 27.3 MPH, with the 25th and 75th percentiles at 22.9 MPH and 33.1 MPH, respectively. These observations indicate a strong association between lower vehicle speed and an increased likelihood of yielding behavior, as reflected in consistently lower speed distributions across all percentiles. 
(2) \textbf{opposite direction yield}: 
the driver yielding rate was 75\% when an oncoming vehicle in the opposite direction yielded, compared to only 3\% when the oncoming vehicle did not yield.
(3) \textbf{speed limit}: roadways with a 30 mph speed limit exhibited a higher yielding rate of 44\% compared to 25\% in 35 mph zones.

Within the category of road networks and infrastructure, all subcategories exert a significant influence on driver yielding behavior:
(1) \textbf{road characteristics}: 
wider roads are associated with increased yielding incidents. The yielding rate at locations without bike lanes (44\%) exceeds that at locations with bike lanes (8\%). Crossings with pedestrian signs exhibit a higher yielding rate (64\%) compared to those without signs (30\%). Similarly, crossings with standard markings of two solid white lines show a yielding rate of 69\%, compared to 10\% at unmarked crossings and 29\% at crossings with continental markings. Furthermore, an increase in the number of bus stops is correlated with a higher likelihood of drivers yielding.
(2) \textbf{distance from public facilities}: proximity to parks is positively associated with an increase in yielding events. 
(3) \textbf{presence of buildings}: the presence of surrounding land uses such as single family homes (41\%), apartments (43\%), commercial buildings (49\%), gas stations or convenience stores (64\%), restaurants or bars (47\%), and parking lots (51\%) is associated with a higher yielding events. In contrast, the absence of the above features, i.e., single family homes (11\%), apartments (9\%), commercial buildings (6\%), gas stations or convenience stores (18\%), restaurants or bars (6\%), and parking lots (16\%), shows markedly lower probability of yielding.

In the category of pedestrian mobility and interaction, two variables are of significant importance: (1) \textbf{number of pedestrians}: drivers exhibit a higher yielding rate of 51\% for more than two pedestrians compared to a yielding rate of 35\% for a pedestrian. (2) \textbf{pedestrian type}: drivers exhibit the highest yielding rate (53.92\%) for pedestrians with strollers or small children, while showing the lowest yielding rate (16.67\%-25.00\%) for non-standard pedestrians such as bicycle riders, scooter/hoverboard users, and persons walking bikes.

\begin{table}[!ht]
    \belowrulesep=0pt
    \aboverulesep=0pt
    \centering
    \caption{Descriptive statistics of adopted categorical variables across yielding and non-yielding events.}
    \renewcommand{\arraystretch}{1.0}
    \resizebox{\textwidth}{!}{
    \begin{tabular}{cc|cccccccc}
    \hline
    \multicolumn{2}{c|}{ \multirow{2}{*}{\pmb{Vehicle Dynamics and Control}} }& \multicolumn{8}{c}{\pmb{Categories}}\\ 
    \cmidrule(r){3-10}
    ~ & ~ & \multicolumn{2}{c}{None} & \multicolumn{2}{c}{Yield} & \multicolumn{2}{c}{Not yield} & ~ & ~   \\ \hline
    \multirow{2}{*}{Opposite direction yield} & Non-yielding  & \multicolumn{2}{c}{55.23\%} & \multicolumn{2}{c}{24.81\%} & \multicolumn{2}{c}{96.79\%} & ~ & ~   \\ 
    ~ & Yielding  & \multicolumn{2}{c}{44.77\%} & \multicolumn{2}{c}{75.19\%} & \multicolumn{2}{c}{3.21\%} & ~ & ~  \\  \hline
    ~ & ~ & \multicolumn{2}{c}{30} & \multicolumn{2}{c}{35} & ~ & ~ & ~ & ~   \\  \hline
    \multirow{2}{*}{Speed limit (MPH)} & Non-yielding  & \multicolumn{2}{c}{56.22\%} & \multicolumn{2}{c}{74.63\%} & ~ & ~ & ~ & ~  \\ 
    ~ & Yielding  & \multicolumn{2}{c}{43.78\%} & \multicolumn{2}{c}{25.37\%} & ~ & ~ & ~ & ~ \\ \hline
    \multicolumn{2}{c|}{ \multirow{2}{*}{ \pmb{Road Networks and Infrastructure}} }& \multicolumn{8}{c}{\pmb{Categories}}\\ 
    \cmidrule(r){3-10}
    ~ & ~ & \multicolumn{2}{c}{Unmarked} & \multicolumn{2}{c}{Standard} & \multicolumn{2}{c}{Continental} & ~ & ~  \\  \hline
    \multirow{2}{*}{Markings} & Non-yielding  & \multicolumn{2}{c}{90.26\%} & \multicolumn{2}{c}{31.04\%} & \multicolumn{2}{c}{70.70\%} & ~ & ~ \\ 
    ~ & Yielding  & \multicolumn{2}{c}{9.74\%} & \multicolumn{2}{c}{68.96\%} & \multicolumn{2}{c}{29.30\%} & ~ & ~   \\  \hline
    ~ & ~ & \multicolumn{2}{c}{No parking} & \multicolumn{2}{c}{One-sided parking} & \multicolumn{2}{c}{Two-sided parking} & ~ & ~  \\ \hline
    \multirow{2}{*}{Presence of on street parking}  & Non-yielding  & \multicolumn{2}{c}{76.58\%} & \multicolumn{2}{c}{83.81\%} & \multicolumn{2}{c}{54.03\%} & ~ & ~  \\ 
    ~ & Yielding  & \multicolumn{2}{c}{23.42\%} & \multicolumn{2}{c}{16.19\%} & \multicolumn{2}{c}{45.97\%} & ~ & ~   \\ \hline
    ~ & ~ & \multicolumn{2}{c}{No present} & \multicolumn{2}{c}{Present} & ~ & ~ & ~ & ~ \\ \hline
    \multirow{2}{*}{Presence of single family} & Non-yielding  & \multicolumn{2}{c}{88.64\%} & \multicolumn{2}{c}{58.80\%} & ~ & ~ & ~ & ~  \\ 
    ~ & Yielding  & \multicolumn{2}{c}{11.36\%} & \multicolumn{2}{c}{41.20\%} & ~ & ~ & ~ & ~  \\  \hline
    ~ & ~ & \multicolumn{2}{c}{No present} & \multicolumn{2}{c}{Present} & ~ & ~ & ~ & ~   \\ \hline
    \multirow{2}{*}{Presence of apartments} & Non-yielding  & \multicolumn{2}{c}{91.06\%} & \multicolumn{2}{c}{57.23\%} & ~ & ~ & ~ & ~  \\ 
    ~ & Yielding  & \multicolumn{2}{c}{8.94\%} & \multicolumn{2}{c}{42.77\%} & ~ & ~ & ~ & ~   \\ \hline
    ~ & ~ & \multicolumn{2}{c}{No present} & \multicolumn{2}{c}{Present} & ~ & ~ & ~ & ~  \\ \hline
    \multirow{2}{*}{Presence of commercial buildings} & Non-yielding  & \multicolumn{2}{c}{94.03\%} & \multicolumn{2}{c}{51.11\%} & ~ & ~ & ~ & ~   \\ 
    ~ & Yielding  & \multicolumn{2}{c}{5.97\%} & \multicolumn{2}{c}{48.89\%} & ~ & ~ & ~ & ~ \\  \hline
    ~ & ~ & \multicolumn{2}{c}{No present} & \multicolumn{2}{c}{Present} & ~ & ~ & ~ & ~  \\ \hline
    \multirow{2}{*}{Presence of gas station/convenient store} & Non-yielding  & \multicolumn{2}{c}{82.06\%} & \multicolumn{2}{c}{35.56\%} & ~ & ~ & ~ & ~   \\ 
    ~ & Yielding  & \multicolumn{2}{c}{17.94\%} & \multicolumn{2}{c}{64.44\%} & ~ & ~ & ~ & ~  \\  \hline
    ~ & ~ & \multicolumn{2}{c}{No present} & \multicolumn{2}{c}{Present} & ~ & ~ & ~ & ~  \\  \hline
    \multirow{2}{*}{Presence of restaurants/bars} & Non-yielding  & \multicolumn{2}{c}{94.42\%} & \multicolumn{2}{c}{52.57\%} & ~ & ~ & ~ & ~   \\ 
    ~ & Yielding  & \multicolumn{2}{c}{5.58\%} & \multicolumn{2}{c}{47.43\%} & ~ & ~ & ~ & ~  \\  \hline
    ~ & ~ & \multicolumn{2}{c}{No present} & \multicolumn{2}{c}{Present} & ~ & ~ & ~ & ~   \\  \hline
    \multirow{2}{*}{Presence of parking lots} & Non-yielding  & \multicolumn{2}{c}{84.04\%} & \multicolumn{2}{c}{49.42\%} & ~ & ~ & ~ & ~   \\ 
    ~ & Yielding  & \multicolumn{2}{c}{15.96\%} & \multicolumn{2}{c}{50.58\%} & ~ & ~ & ~ & ~  \\  \hline
    ~ & ~ & \multicolumn{2}{c}{No present} & \multicolumn{2}{c}{Present} & ~ & ~ & ~ & ~  \\  \hline
    \multirow{2}{*}{Bike lane} & Non-yielding  & \multicolumn{2}{c}{56.35\%} & \multicolumn{2}{c}{91.78\%} & ~ & ~ & ~ & ~   \\ 
    ~ & Yielding  & \multicolumn{2}{c}{43.65\%} & \multicolumn{2}{c}{8.22\%} & ~ & ~ & ~ & ~  \\  \hline
    ~ & ~ & \multicolumn{2}{c}{No present} & \multicolumn{2}{c}{Present} & ~ & ~ & ~ & ~   \\  \hline
    \multirow{2}{*}{Signage} & Non-yielding  & \multicolumn{2}{c}{69.40\%} & \multicolumn{2}{c}{35.94\%} & ~ & ~ & ~ & ~   \\ 
    ~ & Yielding  & \multicolumn{2}{c}{30.60\%} & \multicolumn{2}{c}{64.06\%} & ~ & ~ & ~ & ~  \\ \hline
    \multicolumn{2}{c|}{ \multirow{2}{*}{ \pmb{Pedestrian Mobility and Interaction}} }& \multicolumn{8}{c}{\pmb{Categories}}\\ 
    \cmidrule(r){3-10}
    ~ & ~ & Foot & Bike & Vehicle & Walking bike & With a dog & Stroller/Child & Mix & Other   \\  \hline
    \multirow{2}{*}{Pedestrian type} & Non-yielding  & 53.77\% & 83.10\% & 78.95\% & 83.33\% & 68.10\% & 46.08\% & 53.66\% & 75.00\% \\ 
    ~ & Yielding  & 46.23\% & 16.90\% & 21.05\% & 16.67\% & 31.90\% & 53.92\% & 46.34\% & 25.00\% \\ \hline
    \multicolumn{10}{c}{\begin{tabular}{p{25cm}}
    {Note: pedestrian types are defined as follows. Foot: pedestrian on foot. Bike: pedestrian riding a bicycle. Vehicle: pedestrian using a vehicle, e.g., a scooter or hoverboard. Walking bike: pedestrian walking a bicycle. With a dog: pedestrian accompanied by a dog. Stroller/Child: pedestrian with a stroller or small child. Mix: mixed pedestrian types.}
    \end{tabular}
    }\\ 
    \end{tabular}
    }
    \label{Crosstabs of the summary categorical statistics between yield and non-yield events}
\end{table}

\begin{table}[!ht]
    \belowrulesep=0pt
    \aboverulesep=0pt
    \tiny
    \centering
    \caption{Descriptive statistics of adopted numerical variables across yielding and non-yielding events.}
    \renewcommand{\arraystretch}{1.0}
    \resizebox{\textwidth}{!}{
    \begin{tabular}{cc|ccccc}
    \hline
        \multicolumn{2}{c|}{\multirow{2}{*}{\pmb{Vehicle Dynamics and Control}}}& \multicolumn{5}{c}{\pmb{Value}}  \\
        \cmidrule(r){3-7}
        ~ & ~ & min & 25\% & mean & 75\% & max \\ \hline
        \multirow{2}{*}{Vehicle speed} & Non-yielding (mph) & 1.4 & 22.86 & 27.326 & 33.1 & 90.60  \\ 
        ~ & Yielding (mph) & 0.9 & 5.60 & 10.067 & 12.6 & 50.50  \\ \hline
        \multicolumn{2}{c|}{\multirow{2}{*}{\pmb{Road Networks and Infrastructure}}}& \multicolumn{5}{c}{\pmb{Value}}  \\ 
        \cmidrule(r){3-7}
        ~ & ~ & min & 25\% & mean & 75\% & max \\ \hline
        \multirow{2}{*}{Crossing width} & Non-yielding (ft) & 36.0 & 48.00 & 49.546 & 51.0 & 63.00  \\ 
        ~ & Yielding (ft) & 38.0 & 50.00 & 49.859 & 51.0 & 63.00  \\ \hline
        \multirow{2}{*}{Dist. to nearest school} & Non-yielding (miles) & 0.2 & 0.25 & 0.482 & 0.6 & 4.40  \\ 
        ~ & Yielding (miles) & 0.2 & 0.30 & 0.537 & 0.8 & 4.40  \\ \hline
        \multirow{2}{*}{Dist. to nearest park} & Non-yielding (miles) & 0.1 & 0.20 & 0.282 & 0.4 & 0.53  \\ 
        ~ & Yielding (miles) & 0.1 & 0.20 & 0.202 & 0.4 & 0.53  \\ \hline
        \multirow{2}{*}{Number of bus stops} & Non-yielding & 0 & 0 & 2.143 & 3.0 & 6.00  \\ 
        ~ & Yielding & 0 & 0 & 2.917 & 6.0 & 6.00  \\ \hline
        \multicolumn{2}{c|}{\multirow{2}{*}{\pmb{Pedestrian Mobility and Interaction}}}& \multicolumn{5}{c}{\pmb{Value}}  \\
        \cmidrule(r){3-7}
        ~ & ~ & min & 25\% & mean & 75\% & max \\ \hline
        \multirow{2}{*}{Number of pedestrians} & Non-yielding & 1.0 & 1.00 & 1.391 & 2.0 & 11.00  \\ 
        ~ & Yielding & 1.0 & 1.00 & 1.639 & 2.0 & 11.00  \\ \hline

    \end{tabular}
    }
    \label{Crosstabs of the summary numerical statistics between yield and non-yield events}
\end{table}

\subsection{Working procedures}

This study firstly adopts the data pre-processing pipeline illustrated in Fig.~\ref{fig:Data processing} to partition the input dataset into training and test sets. 
The training set is utilized to identify the most effective traditional classifier among SVM, RF, NN, and LR, with emphasis on achieving good predictive accuracy. Notably, an LR model is developed using stepwise regression to select significant variables for training these classifiers. If LR is determined to be the best-performing baseline model, interpretability will be demonstrated through its coefficients. Otherwise, explainable machine learning techniques will be employed to provide interpretability and facilitate comparison with the LLM-based model.
Additionally, two representative instances from the training set are extracted as few-shot exemplars for the prompt-based evaluation of LLMs. The test set is utilized to evaluate the predictive performance of both the benchmark models and the proposed method, thereby enabling a comparative assessment across modeling paradigms.

Secondly, the descriptive statistics and the best traditional classifier results are incorporated as domain knowledge into the proposed framework in Section \ref{Relevant domain knowledge}. 
The outcomes are further evaluated through comparison with previous literature and assessed for consistency with established theoretical hypotheses.
Finally, this paper evaluates the predictive performance between the best baseline model and the proposed LLM-based approach, and discusses the advantages and limitations of each model, along with their suitable application scenarios. Specifically, this paper selects state-of-the-art LLMs such as GPT-4o, GPT-4o-mini, Deepseek-V3, and Deepseek-R1 as the core of the proposed framework, each accessed via their respective APIs.

\section{Results from traditional classifier} \label{Results from Traditional Classifier}

This paper considers four candidate models for comparison with the proposed LLM-based method.
LR is a statistical model for binary classification that estimates the probability of a binary outcome by fitting data to a logistic curve, producing interpretable results through a linear combination of input features~\citep{schober2021logistic}. 
For binary outcome $Y \in \{0,1\}$ where $Y=1$ for driver yielding and  $Y=0$ for non-yielding, LR models the log-odds of the probability $p=P(Y=1 \mid X)$ by:

\begin{flalign}
\qquad \qquad \log \left(\frac{p}{1-p}\right)=\beta_{0}+\beta_{1} x_{1}+\beta_{2} x_{2}+\cdots+\beta_{n} x_{n}&&
\label{eq:odds}
\end{flalign}
where $\beta_{0}$ is a constant, $\beta_{n}$ is the coefficient of the explanatory variable, $x_{n}$ is the predictor variable, and $n$ is the number of variables. The odds ratio $e^{\beta_{j}}$ indicates the change in odds of the outcome for a one-unit increase in variable $j$. The probability of yielding is given by: 

\begin{flalign}
\qquad \qquad P(Y=1 \mid X)=\frac{\exp \left(\beta_{0}+\beta_{1} x_{1}+\beta_{2} x_{2}+\cdots+\beta_{n} x_{n}\right)}{1+\exp \left(\beta_{0}+\beta_{1} x_{1}+\beta_{2} x_{2}+\cdots+\beta_{n} x_{n}\right)}&&
\end{flalign}

\noindent And the probability of non-yielding is expressed by:
\begin{flalign}
\qquad \qquad P(Y=0 \mid X)=\frac{1}{1+\exp \left(\beta_{0}+\beta_{1} x_{1}+\beta_{2} x_{2}+\cdots+\beta_{n} x_{n}\right)}&&
\end{flalign}

SVM is a binary classification model that maps feature vectors of instances to points in a high-dimensional space~\citep{abdullah2021machine}. The objective of SVM is to identify an optimal hyperplane that maximizes the margin of separation between two classes. An optimal hyperplane can be found by minimizing: 
\begin{flalign}
\qquad \qquad \min _{\mathbf{w}, b} \frac{1}{2}\|\mathbf{w}\|^{2} \quad \text { s.t. } y_{i}\left(\mathbf{w}^{\top} \mathbf{x}_{i}+b\right) \geq 1, i= 1, \cdots, n.&&
\end{flalign}
where $\mathbf{w}$ is the normal vector of the hyperplane, $b$ is the bias term, $\mathbf{x}_{i}$ denotes the feature vector, $y_{i}$ represents the class label, and $n$ is the number of samples. The constraint ensures that each sample lies on the correct side of the margin boundary, achieving maximum margin separation.

The RF model constructs an ensemble of decision trees to enhance both predictive accuracy and generalization~\citep{ferry2024trained}. Each tree is independently trained through random feature selection and bootstrapping, and the final prediction $\widehat{y}$ is obtained by majority voting in classification tasks by
\begin{flalign}
\qquad \qquad \widehat{y}=\text { majority vote }\left(f_{1}(x), \dots, f_i(x), \dots, f_{n}(x)\right) &&
\end{flalign}
where $f_{i}(x)$ represents the prediction of the $i$th tree for the input sample $x$, and $n$ is the number of tree ensembles.

NN is a computational model composed of interconnected layers of artificial neurons that learns hierarchical representations of data through nonlinear transformations, enabling complex pattern recognition and function approximation~\citep{zhao2024review}. Numerous neurons are distributed across multiple hidden layers. Each neuron processes inputs from the previous layer and applies an activation function to generate outputs. Given an input $x$, the output of neuron $j$ in layer $l$ is:
\begin{flalign}
\qquad \qquad z_{j}^{(l)}=f\left(\sum_{i=1}^{n^{(l-1)}} w_{i j}^{(l)} z_{i}^{(l-1)}+b_{j}^{(l)}\right) &&
\end{flalign}
where $f$ is an activation function, $w_{i j}^{(l)}$ represents the mapping weight from the $i$th neuron in layer $l-1$ to the $j$th neuron in layer $l$.  The bias of neuron $j$ in layer $l$ is $b_{j}^{(l)}$, and $n^{(l-1)}$ is the number of neurons in layer $l-1$.

Variable selection improves model interpretability and generalization by retaining only the most informative features while reducing overfitting and complexity. This study employs a stepwise LR, integrating forward selection and backward elimination, to systematically identify the most significant predictors from the initial set of 19 variables using the training dataset.
The optimal features are presented in Table~\ref{tab:logit_results} for the subsequent training of traditional classifiers. 
The standard errors quantify the variability of the coefficient estimates, and the corresponding $z$-scores indicate the number of standard deviations each estimate deviates from the mean. 

The associated $p$-values represent the probability of observing such extreme results under the null hypothesis. 
All selected variables in Table~\ref{tab:logit_results} are statistically significant at the 95\% confidence level as indicated by their low $p$-values. And vehicle speed is identified as the most influential variable among these variables due to its highest absolute $z$-score.
The ``effect" column indicates whether the influence of a variable on yielding behavior is positive or negative.

Moreover, the odds ratio is a key component of stepwise LR as it quantifies the strength and direction of the association between each variable and the outcome, thereby enhancing the interpretability of model results~\citep{zabor2022logistic}.
For example, each unit increase in vehicle speed is associated with a $1 - exp(-0.240) = 0.21$ reduction in the relative probability of driver yielding when other variables remain fixed. 
The model also indicates that yielding behavior by drivers in the opposite direction reduces the relative probability of yielding by 80.7\%. In contrast, wider major roads positively influence yielding, with each additional foot of width associated with a 10.3\% increase in the odds of yielding.
The infrastructure environment also shows substantial effects on drive yielding. The presence of restaurants or bars is associated with an odds ratio of 6.24, corresponding to a 524\% increase in the relative probability of yielding compared to sites without such facilities. The presence of parking lots corresponds to an odds ratio of 0.197, indicating a reduction in yielding probability. 
Additionally, proximity to parks and schools exhibits a positive influence on yielding behavior, with each mile closer to a park associated with a 98\% increase in the probability of yielding and each mile closer to a school associated with a 36\% increase.

It should be noted that the discrepancy observed between the odds ratios of ``opposite direction yield”, ``presence of parking lots”, and ``distance to the nearest school” in LR and their corresponding patterns in the descriptive statistics of Section \ref{Descriptive Statistics} can be attributed to confounding effects.
%
In LR, the odds ratio for a given predictor represents its net effect on the outcome, conditional on all other variables fixed.
In contrast, descriptive statistics capture only the marginal association between a variable and the outcome, which may be influenced by uncontrolled factors~\citep{pourhoseingholi2012control,Greenland_2016}.
Therefore, odds ratios derived from LR offer a more robust and causally interpretable assessment of variable effects compared to simple descriptive comparisons, and the results from LR are adopted as the primary reference in this study whenever discrepancies arise.

\begin{table}[h!]
\centering
\caption{Significant variables identified through the stepwise LR.}
\small  
\setlength{\tabcolsep}{4pt}  
\begin{tabular}{lccccc}
\hline
\textbf{Variable} & \textbf{Coefficient} & \textbf{Standard error} & $\boldsymbol{z}$-\textbf{score} & $\boldsymbol{p}$-\textbf{value} & \textbf{Effect} \\
\hline
\text{Crossing width} & 0.098 & 0.010 & 9.38 & 0.000 & + \\
\text{Presence of restaurants/bars} & 1.832 & 0.409 & 4.48 & 0.000 & + \\ 
\text{Vehicle speed} & -0.240 & 0.011 & -20.99 & 0.000 & - \\
\text{Distance to nearest school} & -0.446 & 0.163 & -2.74 & 0.006 & - \\
\text{Presence of parking lots} & -1.624 & 0.263 & -6.18 & 0.000 & - \\
\text{Opposite direction yield} & -1.644 & 0.225 & -7.29 & 0.000 & - \\
\text{Distance to nearest park} & -4.095 & 0.996 & -4.11 & 0.000 & - \\
\hline
\end{tabular}
\label{tab:logit_results}
\end{table}

\begin{figure*}[!htb]
\centering
\includegraphics[width = 0.5\columnwidth]{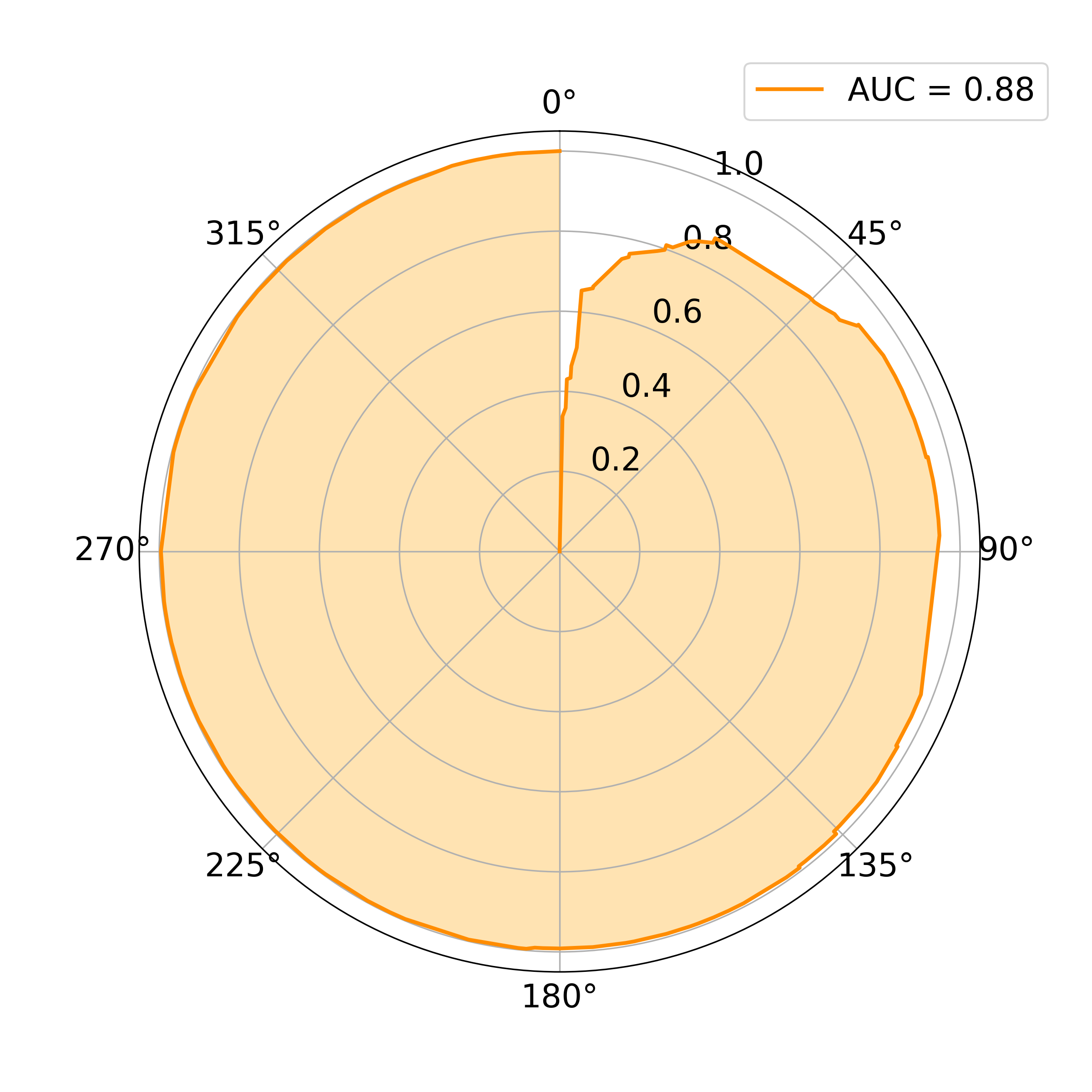}
\caption{Polar ROC curves for the LR model.} 
\label{ROC}
\end{figure*}

Fig.~\ref{ROC} presents the receiver operating characteristic (ROC) curve for LR, which illustrates the trade-off between the true positive rate and the false positive rate across various decision thresholds. 
The area under the ROC curve (AUC) reaches 0.88 on the test dataset. An AUC value closer to 1 indicates superior classification performance, as it reflects the model’s enhanced ability to discriminate between positive and negative instances across all thresholds. This result also confirms the relevance of the selected explanatory variables in capturing the underlying dynamics of driver–pedestrian interactions.
Notably, SVM (AUC = 0.88), RF (AUC = 0.885), and NN (AUC = 0.89) achieved comparable predictive performance against LR. 
In this study, LR is employed as the baseline method and as a pedagogical model for instructing the LLM to enhance interpretability.

\section{LLM framework}\label{Methodology}

\begin{figure}[!htb]
\centering
\includegraphics[width = \columnwidth]{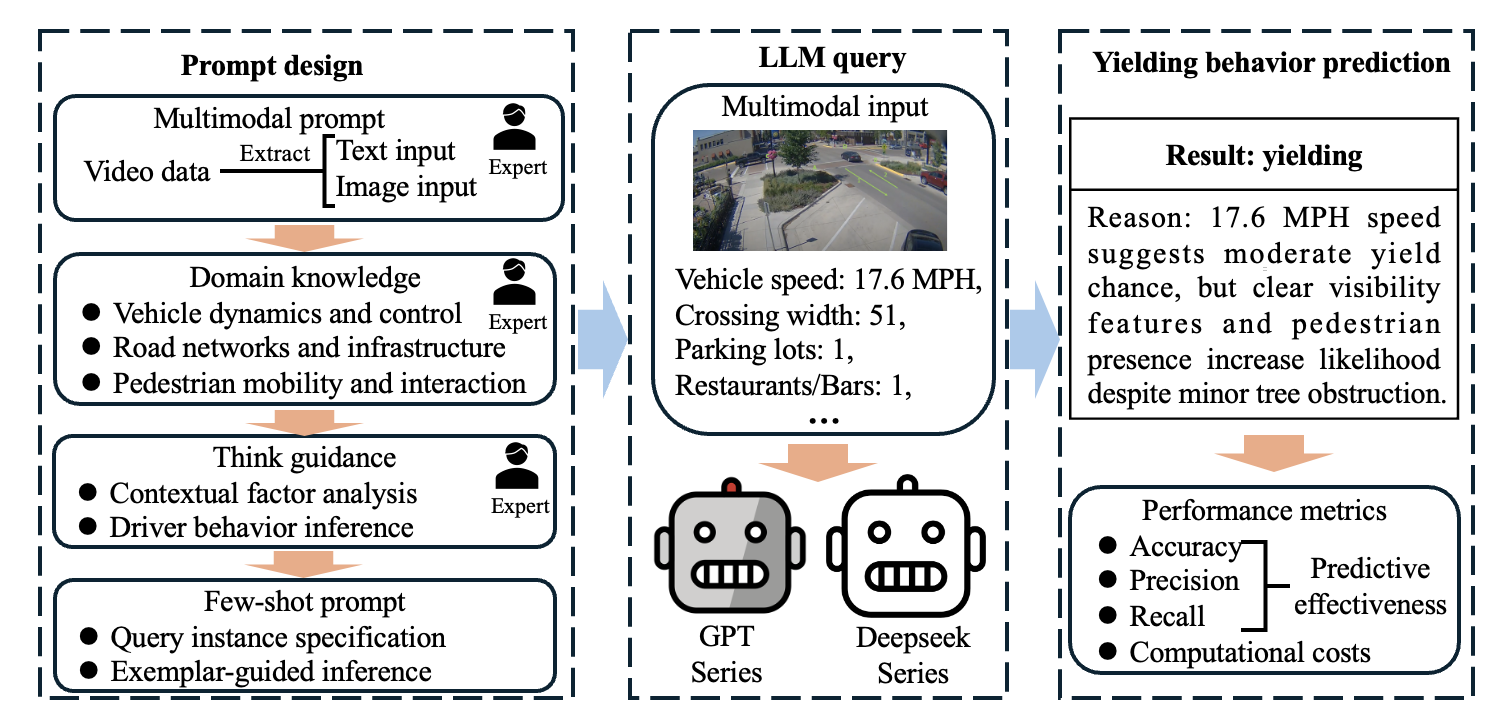}
\caption{LLM framework tailored for driver yielding behavior prediction. The expert symbol indicates that domain expertise is required for a component of the prompt design.} 
\label{Framework}
\end{figure}

LLMs serve as general-purpose multi-tasking models capable of predicting driver yielding behavior, and incorporating detailed contextual information significantly enhances their predictive performance~\citep{Mo2023}.
This section introduces the proposed framework with three key stages for the identification of driver yielding behavior in Fig.~\ref{Framework}. 

In the first stage, textual data extracted from videos is integrated with domain knowledge, structured thinking guidance, and few-shot exemplars during the prompt design. 
In the second stage, a multimodal prompt with site specific images forms the comprehensive inputs for various state-of-the-art LLMs to infer yielding behavior over the heterogeneous traffic data. 
Finally, each adopted LLM yields a binary yielding decision with an interpretable explanation of its reasoning process. The performance of each model is assessed using accuracy, precision, and recall, enabling a comparative assessment across the different LLM platforms. Further details of the prompt design phase are provided in the following subsections.

\subsection{Relevant domain knowledge} \label{Relevant domain knowledge}
    
\begin{figure}[!htb]
    \centering
    \includegraphics[width = \columnwidth]{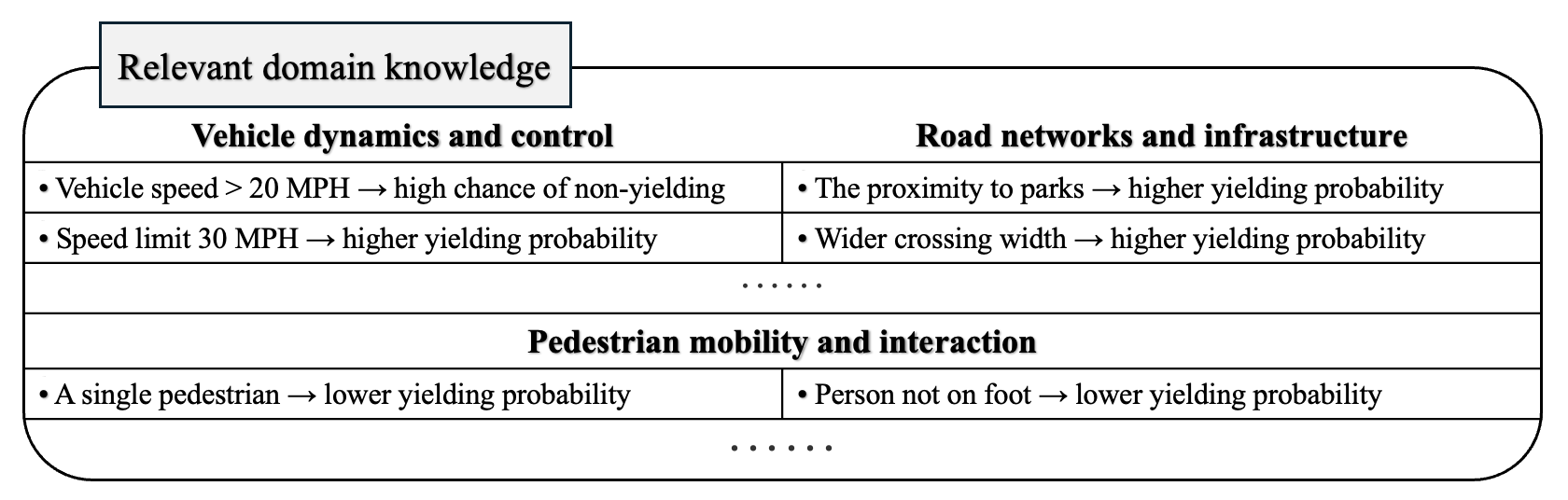}
    \caption{Selected key elements of domain specific knowledge embedded into the prompt design.} 
    \label{Relevant domain knowledge_image}
    \end{figure}

LLMs are general-purpose models trained on vast amounts of data, making them highly versatile across various applications. However, their performance in domain-specific prediction tasks is often suboptimal~\citep{Wandelt2024}. A common approach to enhancing LLMs’ performance in specialized fields is fine tuning with domain specific data. However, the fine tuning process typically requires extensive labeled datasets and substantial computational resources, making it costly and resource intensive. 
To address this challenge, domain-specific knowledge is incorporated into the prompt in textual form as shown in Fig.~\ref{Relevant domain knowledge_image}, enabling LLMs to utilize relevant expertise for reasoning without additional model training.

The domain knowledge integrated into our framework is mainly derived from the descriptive statistics in Section~\ref{Descriptive Statistics} and the results of the stepwise LR in Section~\ref{Results from Traditional Classifier}. This knowledge in Fig.~\ref{Relevant domain knowledge_image} is organized in an object-oriented manner and categorized into three core dimensions.
It should be noted that results obtained from stepwise LR are considered as the primary reference for informing domain knowledge whenever they diverge from the findings of descriptive statistics.
In the category of vehicle dynamics and control, each one-mile-per-hour increase in vehicle speed is associated with a 0.79-fold decrease in the odds of yielding, consistent with findings in prior traffic behavior research~\citep{Silvano2016second}.
Descriptive statistics indicate that the yielding rate in 30 mph zones was 19\% higher than in 35 mph zones, aligning with empirical observations in the field~\citep{bertulis2014}.
Within the category of road networks and infrastructure, the proximity to parks or schools is significantly associated with increased driver-yielding rates. Specifically, each one-mile decrease in distance to the nearest park corresponds to a 98\% increase in the probability of yielding, and each one-mile decrease in distance to the nearest school is associated with a 36\% increase in yielding probability.
An increase in pedestrian numbers is positively correlated with the rate of driver yielding behavior.
Furthermore, wider intersection width was positively associated with yielding probability on a per-foot basis, consistent with~\cite{Schroeder2011}.

For brevity, other domain knowledge is not detailed herein, and it aligns with perspectives established in previous literature.
Guiding the model with these structured, domain-informed reasoning frameworks shall enhance its ability to generate reliable predictions in specialized traffic safety scenarios while minimizing the need for computationally intensive fine-tuning.

     \subsection{Thinking guidance} \label{Thinking guidance}

\begin{figure}[!htb]
    \centering
    \includegraphics[width = \columnwidth]{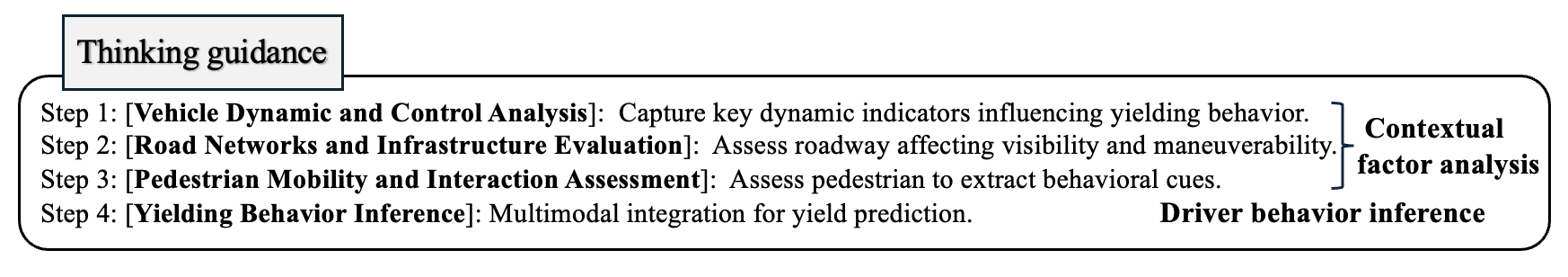}
    \caption{Procedures of thinking guidance adopted in the prompt design.} 
    \label{Simplified thinking guidance}
    \end{figure}

Prompting strategies such as chain-of-thought and plan-and-solve have substantially improved LLM performance by decomposing complex tasks into structured reasoning steps~\citep{wei2022Cot, wang2023PS}. Recently, meta-prompting has further advanced this approach by prioritizing structural and syntactic task representations over content-level details, enabling more abstract and systematic model reasoning~\citep{zhang2023meta}.
However, LLM-generated text is often unstructured and may include excessive or irrelevant content, reducing the effectiveness of the predictions. 
To enhance prediction quality, thinking guidance in Fig.~\ref{Simplified thinking guidance} is incorporated into the prompt design to establish a structured reasoning process for driver behavior prediction, resulting in a final binary classification of whether the driver yields.
Compared to chain-of-thought and meta-prompting strategies, structured thinking guidance provides greater domain adaptability, a clearer logical structure, and improved controllability. Its alignment with the characteristics of task-specific data makes it particularly well-suited for applications such as traffic safety prediction, where fine-grained causal reasoning is essential.

The proposed thinking guidance reduces extraneous output and enhances the robustness and interpretability of model reasoning. In Fig.~\ref{Simplified thinking guidance}, the thinking guidance integrates analyses from three dimensions related to vehicle attributes, road infrastructure, and pedestrian interactions, to deliver a final decision with concise justification.
Unlike domain knowledge, which serves as an encyclopedic repository containing factual information, thinking guidance operates as a procedural framework that specifies how this knowledge is accessed, organized, and applied during reasoning. Namely, domain knowledge specifies the ``what”, whereas thinking guidance defines the ``how”, guiding LLMs to appropriately leverage domain-specific information in a step-by-step manner to support coherent inference and problem-solving. We will introduce key steps of think guidance as follows.

\noindent \textbf{Step 1: Vehicle Dynamics and Control Analysis}. Vehicle-related factors, e.g., vehicle speed, have been identified as the most significant determinants of yielding behavior.
The LLM integrates its intrinsic reasoning with relevant domain knowledge to comprehensively evaluate vehicle-related attributes, thereby estimating the likelihood of the driver yielding to pedestrians from the vehicle’s perspective.

\noindent \textbf{Step 2: Road Networks and Infrastructure Evaluation}. This step focuses on environmental and infrastructural features that affect driving complexity and visibility. 
Attributes such as crossing width are taken into account, as wider roads are associated with higher yielding rates due to drivers’ perception of increased pedestrian exposure and elevated crossing risk.
%
In this step, the LLMs leverage surrounding road and environmental conditions and integrate internal reasoning with domain-specific knowledge to evaluate the likelihood of driver yielding behavior from the perspective of road conditions.

\noindent \textbf{Step 3: Pedestrian Mobility and Interaction Assessment}. The LLMs examine variables such as pedestrian group size and pedestrian type (e.g., riding a bike or walking). These features offer subtle but important cues about pedestrian intent and vulnerability, which may influence a driver’s decision-making. At this step, the LLM evaluates pedestrian characteristics to infer the likelihood of a driver yielding. 

\noindent \textbf{Step 4: Yielding Behavior Inference}. This step integrates insights from the above three stages. 
By integrating the statistical significance of each factor with their inherent reasoning capabilities, LLMs can generate more accurate predictions while enhancing interpretability.
This structured, multi-step reasoning chain helps bridge the gap between statistical correlation and contextual understanding, allowing for more robust modeling of pedestrian safety at unsignalized intersections.

     \subsection{Few-shot prompting} 
\begin{figure}[!htb]
    \centering
    \includegraphics[width = \columnwidth]{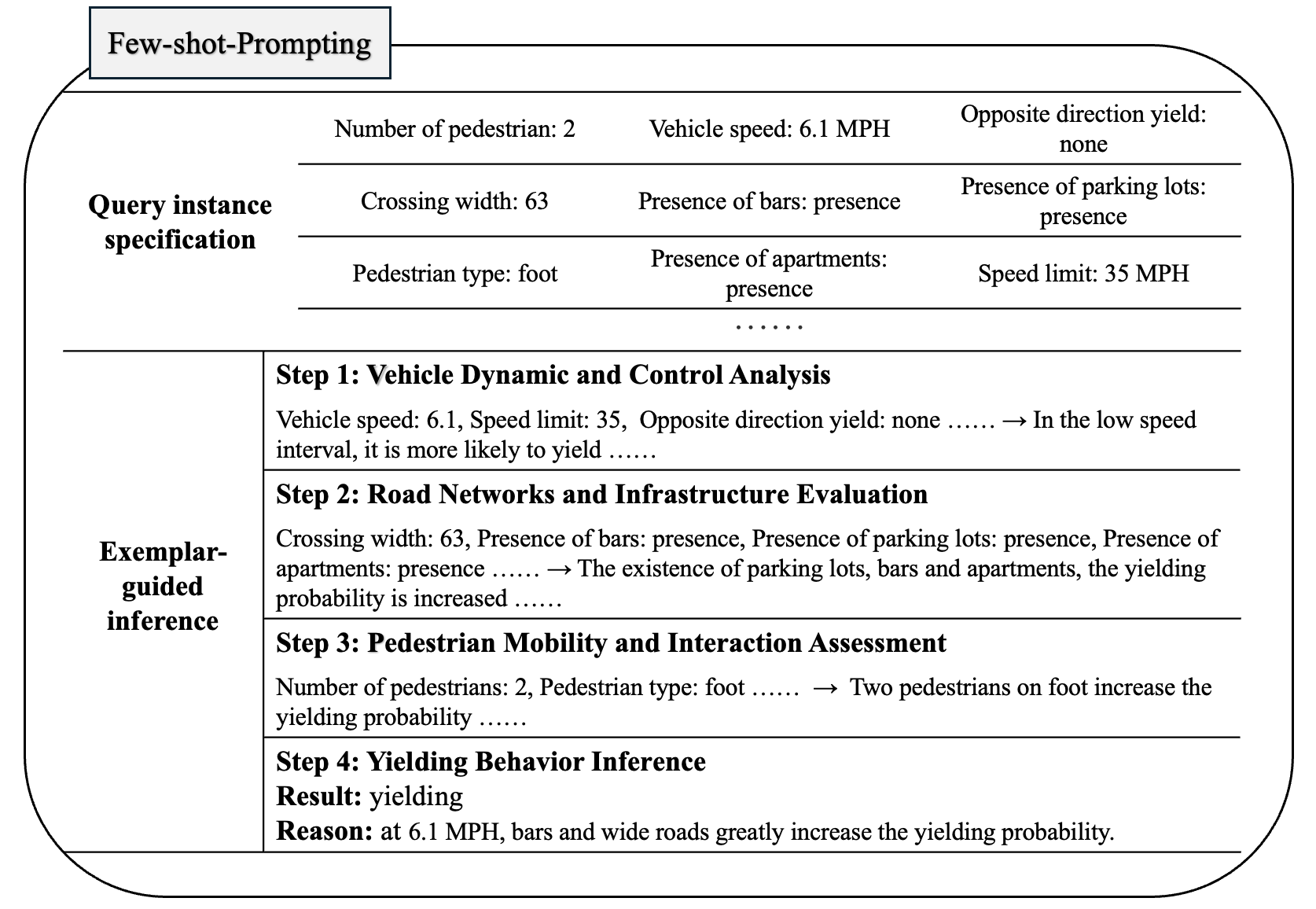}
    \caption{Example of few-shot prompting incorporated into the prompt design.} 
    \label{Few_shot_Prompting}
    \end{figure}

    While LLMs demonstrate remarkable zero-shot capabilities~\citep{wei2021zero}, they often struggle with more complex tasks when operating in a purely zero-shot setting. Few-shot~\citep{zhang2020fewshot} learning serves as an effective technique to enable in context learning, where demonstrations are provided within the prompt to guide the model toward improved performance. To enhance predictive accuracy, the prompt design incorporates a limited number of representative examples. These examples follow the structured reasoning steps outlined in the thinking guidance framework, ensuring that the model adheres to a predefined logical process when generating predictions. By integrating these demonstrations into the prompt, the LLM is better equipped to follow the reasoning path necessary for accurate classification of driver yielding behavior. As illustrated in Fig.~\ref{Few_shot_Prompting}, this approach leverages few-shot prompting to improve model performance by offering contextual references, effectively reducing errors and enhancing consistency. By integrating the structured thinking guidance outlined in Section \ref{Thinking guidance} with real-world examples and independently evaluating the scenario from multiple perspectives, the framework ultimately produces a comprehensive assessment of the likelihood that the driver will yield. The framework ensures more reliable and interpretable predictions while maintaining computational efficiency. 

\subsection{Multimodal prompting} 

\begin{figure}[tp]
\centering
\subfloat[Site 1]{  
	\includegraphics[width=0.3\linewidth]{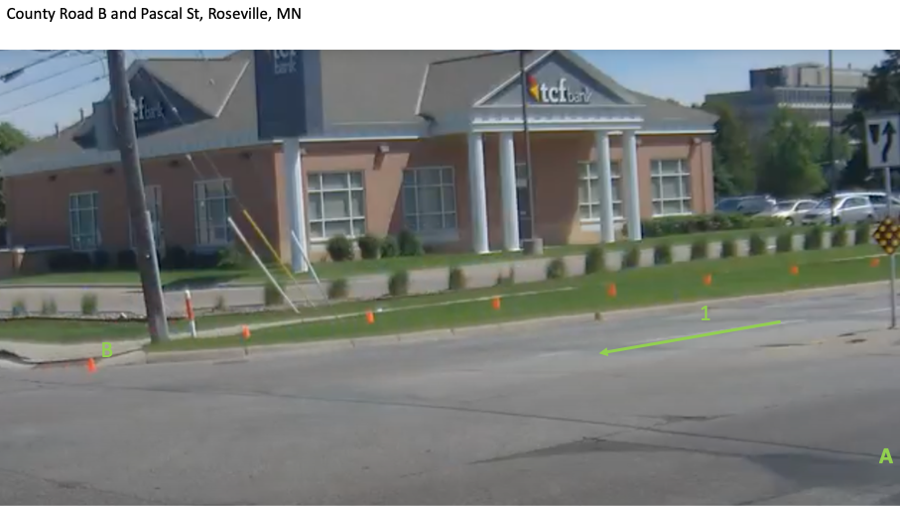}}
\subfloat[Site 2]{
	\includegraphics[width=0.3\linewidth]{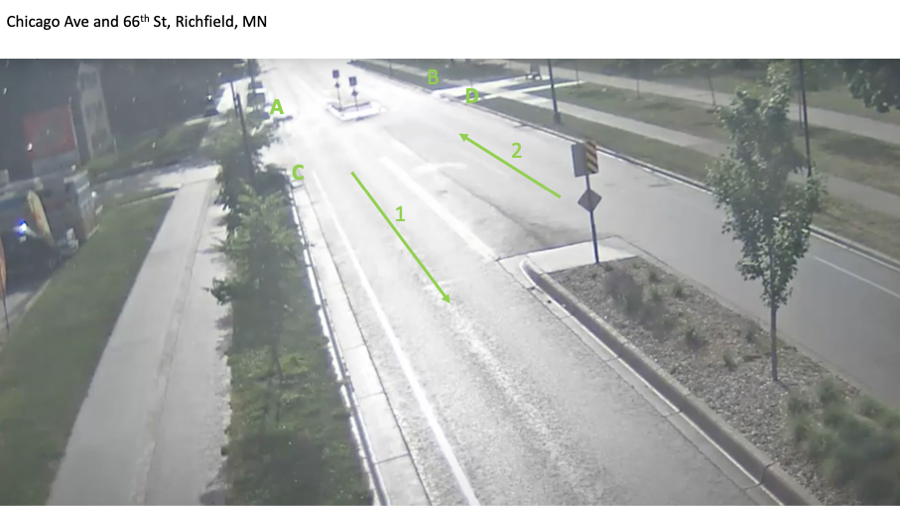}}
\subfloat[Site 3]{
	\includegraphics[width=0.3\linewidth]{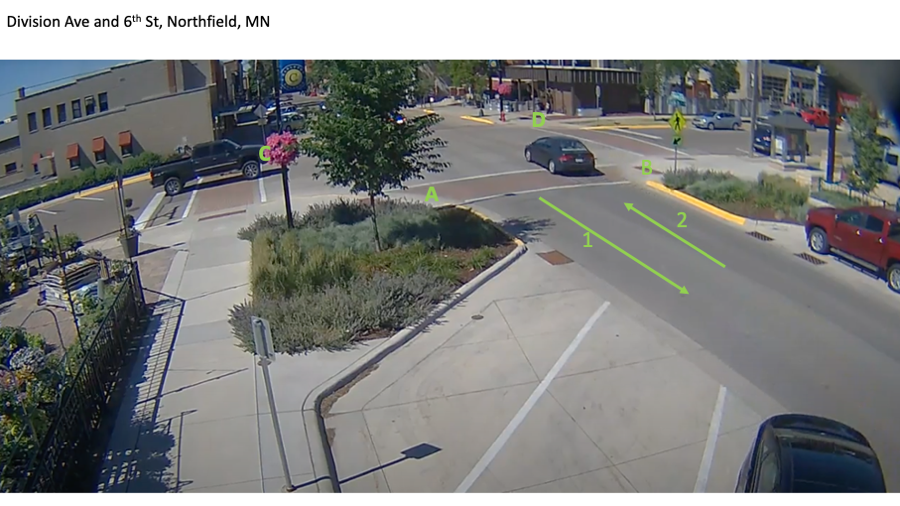}}
\hfill
\subfloat[Site 6]{  
	\includegraphics[width=0.3\linewidth]{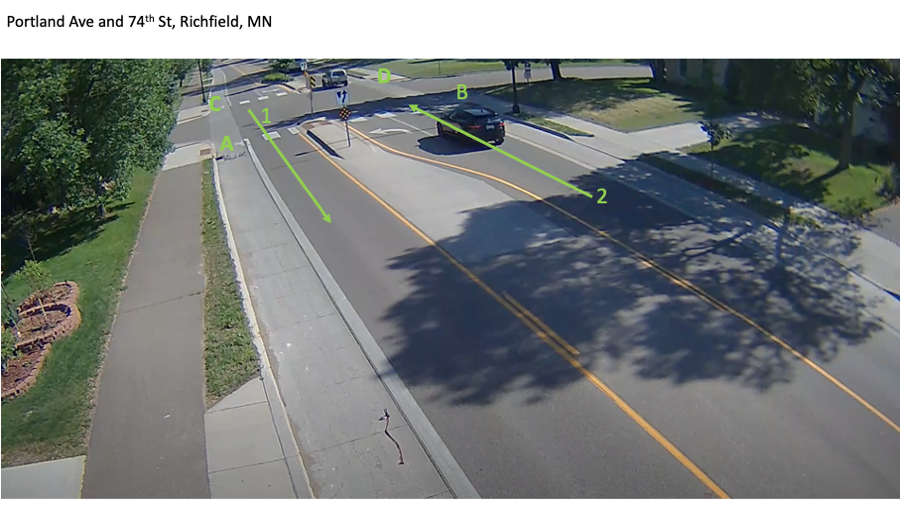}}
\subfloat[Site 7]{
	\includegraphics[width=0.3\linewidth]{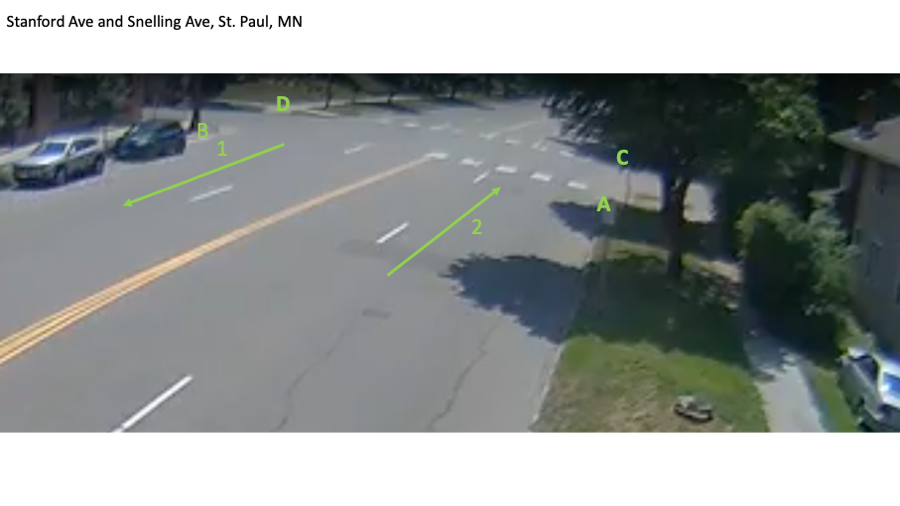}}
\subfloat[Site 13]{
	\includegraphics[width=0.3\linewidth]{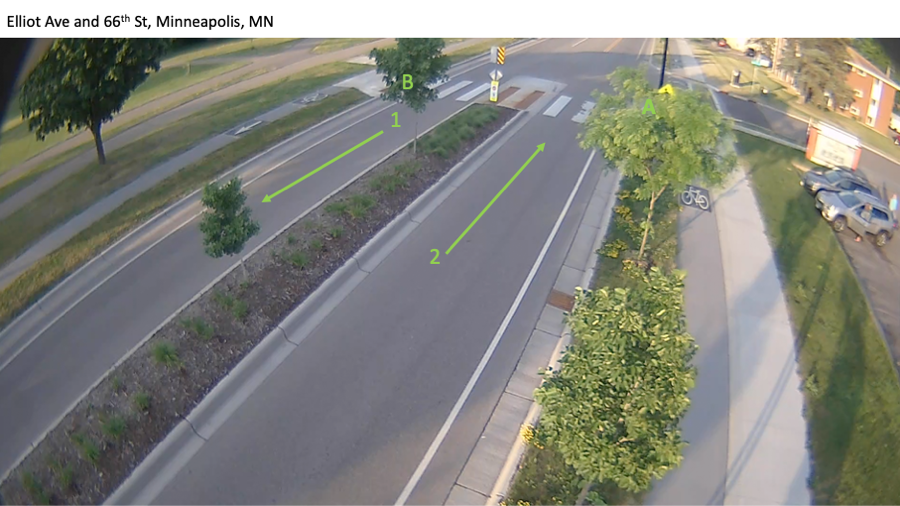}}
\hfill
\subfloat[Site 14]{  
	\includegraphics[width=0.3\linewidth]{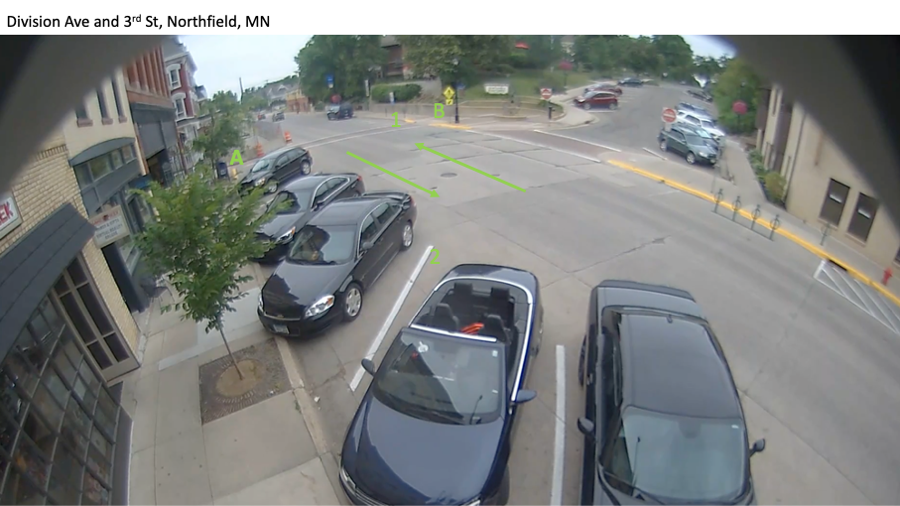}}
\subfloat[Site 16]{
	\includegraphics[width=0.3\linewidth]{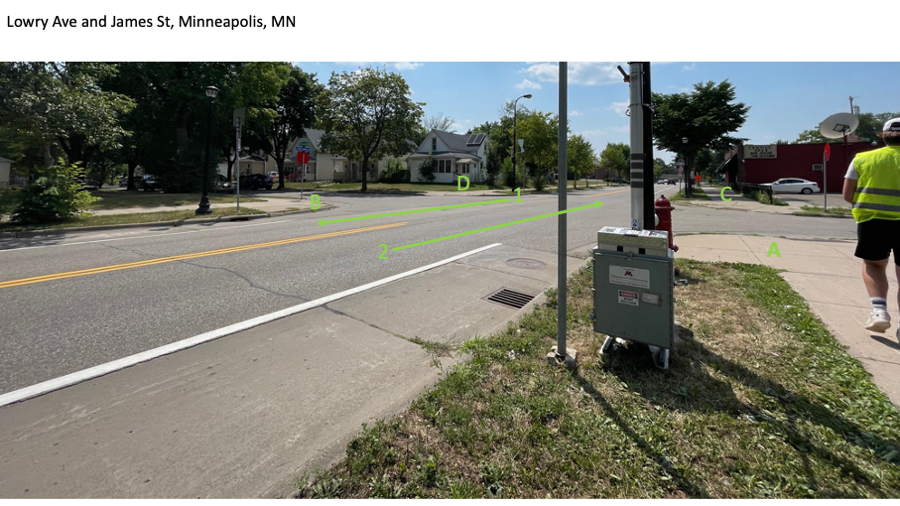}}
\subfloat[Site 18]{
	\includegraphics[width=0.3\linewidth]{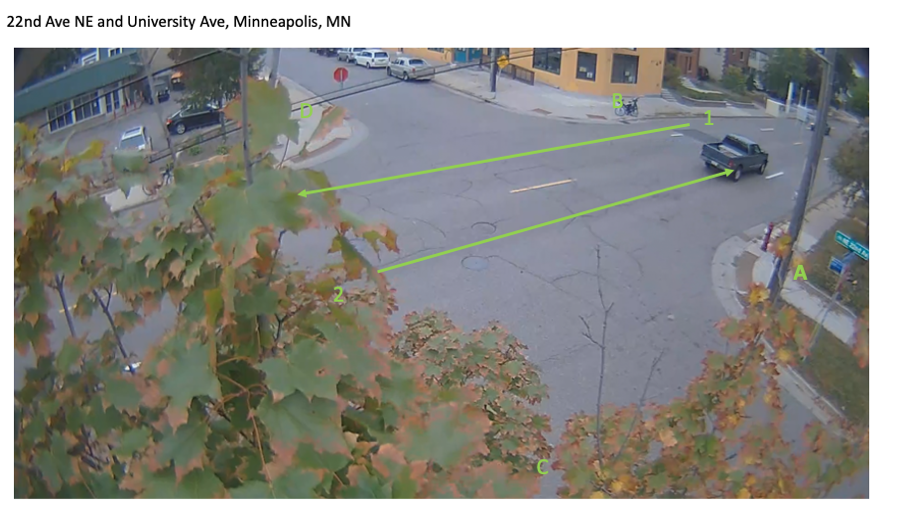}}
\hfill
\caption{Representative surveillance images captured at key monitoring locations.}\label{fig:intersection figures}
\end{figure}

    The raw data used in this study consists of video recordings, which are manually processed into structured tabular data capturing key attributes of vehicles, pedestrians, and intersection features, such as vehicle speed, intersection width, and pedestrian group size. However, despite careful manual processing, certain subtle but important characteristics of intersections may not be fully represented. It includes elements such as visual obstructions, the clarity of traffic signage, and the visibility or condition of lane markings. To solve this problem, we adopted the multimodal LLM method~\citep{zhang2023mul}, which integrates textual data with a representative image from each intersection, as illustrated in Fig.~\ref{fig:intersection figures}.
    In this study, images are not used to teach the LLM in the same sense as domain knowledge or structured reasoning guidance. Instead, they serve as contextual evidence that complements textual and numerical inputs.
    This method is applied to LLMs that support multimodal inputs, enabling the model to simultaneously process structured numerical data and unstructured visual information. These images provide supplementary spatial and visual cues, such as the presence of trees blocking sightlines, crosswalk visibility, and signage conditions. By combining structured inputs with visual representations, the multimodal framework enables the LLM to access a richer spectrum of intersection specific information. It not only enhances the contextual understanding of traffic environments but also improves the model’s ability to generalize and predict driver yielding behavior with greater accuracy and robustness.

\subsection{Handling missing data}

Given that the dataset consists of real-world, naturalistic observations, a small amount of missing or incomplete contextual data is expected. This study does not employ traditional imputation methods to handle missing values. On the contrary, specific statements can be added to the prompt template to enable the LLM to dynamically assess whether missing data should be input or to safely ignore the missing data during the prediction period.

This built-in reasoning refers to the model’s intrinsic ability to infer, contextualize, and adapt its predictions based on incomplete information by leveraging patterns and relationships learned during pre-training. Rather than relying on fixed imputation rules or external preprocessing, the LLM uses its understanding of domain knowledge and contextual cues embedded within the prompt to make informed judgments about the relevance and impact of missing data. This flexible reasoning mechanism allows the model to maintain robustness in the face of real-world data complexities and uncertainties.

\section{Experiment}\label{Experiment}

To evaluate the effectiveness of the proposed framework, we conducted a series of experiments. The experimental setup includes parameter configurations and the selection of a baseline model, followed by the use of well-established performance metrics for comparison. We then present the results of predictive performance, interpretability, and computational cost and time analysis, highlighting both the strengths and limitations of LLM-based approaches in modeling driver yielding behavior.

\subsection{Parameter settings and baseline model}

The language models utilized in our experiments include GPT-4o-mini, GPT-4o, Deepseek-V3, and Deepseek-R1. The prompt templates employed for these models are consistent with the specifications described in Section~\ref{Methodology}. Among them, the multimodal model accepts both image and text inputs, while the others are limited to text based prompts. LR was ultimately chosen as our baseline model as it predicts as well as other classifiers and has high interpretability, and was implemented using the \texttt{sklearn} library in \texttt{Python}.

\subsection{Performance metrics}

This paper adopts three performance metrics for comprehensive evaluation. The precision measures the proportion of true positive predictions among all positive predictions made by the model. It reflects the model's ability to avoid false positives and is defined as:
\begin{flalign}\label{Precision}
\qquad \qquad \text { Precision }=\frac{\text { TP }}{\text { TP }+ \text { FP }}&&
\end{flalign}
where TP is the positive sample predicted by the model as the positive class, TN is the negative sample predicted by the model as the negative class, FP is the negative sample predicted by the model as the positive class, and FN is the positive sample predicted by the model as a negative class.
The recall measures the proportion of true positive predictions among all actual positive cases. It indicates the model’s ability to identify all relevant instances, and is defined as:
\begin{flalign}\label{Recall}
\qquad \qquad \text { Recall }=\frac{\text { TP }}{\text { TP }+ \text { FN }}&&
\end{flalign}
The overall accuracy represents the proportion of all correct predictions (both true positives and true negatives) among the total number of predictions. It provides a general measure of model performance, and is defined as: 
\begin{flalign}\label{Accuracy}
\qquad \qquad \text { Accuracy }=\frac{\text { (TP + TN) }}{\text { (TP + TN + FP + FN) }}&&
\end{flalign}

\subsection{Result}

We conducted a comprehensive evaluation of five models, namely LR, GPT-4o mini, GPT-4o, Deepseek-V3, and Deepseek-R1, in predicting driver yielding behavior across 18 intersections. As summarized in Table~\ref{comparison results}, the results indicate substantial variation in model performance depending on the complexity of the traffic environment, with GPT-4o consistently demonstrating the highest predictive accuracy.

\begin{table}[!ht]
    \belowrulesep=0pt
    \aboverulesep=0pt
    \renewcommand{\arraystretch}{1.4}
    \centering
    \caption{Comparative prediction performance of LR, GPT-4o-mini, GPT-4o, Deepseek-V3, and Deepseek-R1. Note that the best-performing value for each metric is highlighted in bold.}
    \resizebox{\textwidth}{!}{
    \begin{tabular}{c|ccccccccccccccc}
    \hline
        \multirow{2}{*}{Site} & \multicolumn{3}{c}{LogisticRegression} &\multicolumn{3}{c}{GPT-4o-mini} &\multicolumn{3}{c}{GPT-4o} &\multicolumn{3}{c}{Deepseek-V3} &\multicolumn{3}{c}{Deepseek-R1}\\ 
        \cmidrule(r){2-4} \cmidrule(r){5-7} \cmidrule(r){8-10} \cmidrule(r){11-13} \cmidrule(r){14-16}
        ~ & Accuracy & Recall & Precision & Accuracy & Recall & Precision & Accuracy & Recall & Precision & Accuracy & Recall & Precision & Accuracy & Recall & Precision\\ 
        \hline
        Site 1 & $0.8000$  & $0.5000$  & $1.0000$  & $0.8000$  & $0.5000$  & $1.0000$  & $0.8000$  & $0.5000$  & $1.0000$ & $0.6000$  & $0.0000$  & $0.0000$ & $0.8000$  & $0.5000$  & $1.0000$\\ 
        Site 2 & $0.9231$  & $1.0000$  & $0.8750$  & $0.5385$  & $0.1429$  & $1.0000$  & $0.9231$  & $0.8571$  & $1.0000$  & $ 0.6923$  & $0.4286$  & $1.0000$ & $ 1.0000$  & $1.0000$  & $1.0000$\\ 
        Site 3 & $0.8034$  & $0.9036$  & $0.8333$  & $0.5641$  & $0.4819$  & $0.8333$  & $0.8632$  & $1.0000$  & $0.8384$ & $0.6923$  & $0.6506$  & $0.8852$ & $0.6667$  & $0.9277$  & $0.7000$\\ 
        Site 4 & $0.9667$  & $0.5000$  & $1.0000$  & $0.9667$  & $0.5000$  & $1.0000$  & $0.9667$  & $0.5000$  & $1.0000$ & $0.9667$  & $0.5000$  & $1.0000$ &$0.9667$  & $0.5000$  & $1.0000$ \\ 
        Site 5 & $1.0000$  & $1.0000$  & $1.0000$  & $1.0000$  & $1.0000$  & $1.0000$  & $1.0000$  & $1.0000$  & $1.0000$ & $1.0000$  & $1.0000$  & $1.0000$ & $1.0000$  & $1.0000$  & $1.0000$ \\ 
        Site 6 & $0.7857$  & $0.5000$  & $0.6667$  & $0.7143$  & $0.2500$  & $0.5000$  & $0.7143$  & $0.5000$  & $0.5000$ & $0.7143$  & $0.0000$  & $0.0000$ & $0.7143$  & $0.2500$  & $0.5000$\\ 
        Site 7 & $1.0000$  & $1.0000$  & $1.0000$  & $0.9231$  & $0.5000$  & $1.0000$  & $1.0000$  & $1.0000$  & $1.0000$ & $0.9231$  & $0.5000$  & $1.0000$ & $1.0000$  & $1.0000$  & $1.0000$\\ 
        Site 8 & $0.9412$  & $0.5000$  & $1.0000$  & $0.8824$  & $0.0000$  & $0.0000$  & $0.9412$  & $0.5000$  & $1.0000$ & $0.9412$  & $0.5000$  & $1.0000$ & $0.9412$  & $1.0000$  & $0.6667$\\ 
        Site 9 & $1.0000$  & $0.0000$  & $0.0000$  & $1.0000$  & $0.0000$  & $0.0000$  & $1.0000$  & $0.0000$  & $0.0000$ & $1.0000$  & $0.0000$  & $0.0000$ & $1.0000$  & $0.0000$  & $0.0000$\\ 
        Site 10 & $0.9412$  & $0.0000$  & $0.0000$  & $0.9412$  & $0.0000$  & $0.0000$  & $0.9412$  & $0.0000$ & $0.0000$ & $0.9412$  & $0.0000$ & $0.0000$ & $0.9412$  & $0.0000$  & $0.0000$\\ 
        Site 11 & $1.0000$  & $0.0000$  & $0.0000$  & $1.0000$  & $0.0000$  & $0.0000$  & $1.0000$  & $0.0000$  & $0.0000$ & $1.0000$  & $0.0000$  & $0.0000$ & $1.0000$  & $0.0000$  & $0.0000$\\ 
        Site 12 & $0.9500$  & $0.0000$  & $0.0000$  & $0.9500$  & $0.0000$  & $0.0000$  & $0.9500$  & $0.0000$  & $0.0000$ & $0.9500$  & $0.0000$  & $0.0000$ & $0.8500$  & $0.0000$  & $0.0000$\\ 
        Site 13 & $1.0000$  & $1.0000$  & $1.0000$  & $0.4545$  & $0.1429$  & $1.0000$  & $1.0000$  & $1.0000$  & $1.0000$ & $0.8182$  & $0.7143$  & $1.0000$ & $1.0000$  & $1.0000$  & $1.0000$\\ 
        Site 14 & $0.8667$  & $1.0000$  & $0.8519$  & $0.6333$  & $0.6087$  & $0.8750$  & $0.9000$  & $1.0000$  & $0.8846$ &$0.8667$  & $0.9130$  & $0.9130$ &$0.8333$  & $0.9565$  & $0.8462$\\ 
        Site 15 & $1.0000$  & $0.0000$  & $0.0000$  & $0.9000$  & $0.0000$  & $0.0000$  & $0.9000$  & $0.0000$  & $0.0000$ & $1.0000$  & $0.0000$  & $0.0000$ & $0.9000$  & $0.0000$  & $0.0000$\\ 
        Site 16 & $0.8182$  & $0.9495$  & $0.8174$  & $0.5035$  & $0.2929$  & $0.9667$  & $0.8881$  & $0.9192$  & $0.9192$ & $0.8741$  & $0.8788$  & $0.9355$ & $0.8106$  & $0.9495$  & $0.8624$\\ 
        Site 17 & $0.9375$  & $0.0000$  & $0.0000$  & $0.9375$  & $0.0000$  & $0.0000$  & $0.9375$  & $0.0000 $ & $0.0000$ & $0.9375$  & $0.0000$  & $0.0000$ & $0.9375$  & $0.0000$  & $0.0000$\\ 
        Site 18 & $0.9474$  & $0.5000$  & $1.0000$  & $0.8947$  & $0.0000$  & $0.0000$  & $0.9474$  & $0.5000$  & $1.0000$ & $0.8947$  & $0.0000$  & $0.0000$ & $0.8947$  & $0.0000$  & $0.0000$\\ 
        Overall & $0.8808$  & $0.9072$  & $0.8366$  & $0.7020$  & $0.3797$  & $0.8738$  & \pmb{$0.9106$}  & \pmb{$0.9241$}  & $0.8795$ & $0.8547$  & $0.7342$  & \pmb{$ 0.9206$} & $0.8529$  & $0.9072$  & $0.7904$\\ \hline
    \end{tabular}
    }
    \label{comparison results}
\end{table}

Across all sites, GPT-4o emerged as the most balanced and reliable predictor, achieving the highest overall accuracy (91.06\%), recall (92.41\%), and precision (87.95\%). These results confirm its ability to both minimize false positives and maximize true positive detections, making it well-suited for pedestrian safety systems that demand contextual reasoning under dynamic conditions. Deepseek-V3 followed closely with an accuracy of 85.47\%, recall of 73.42\%, and precision of 92.06\%, highlighting its strength in avoiding false positives, critical in applications such as collision avoidance alerts. Deepseek-R1 demonstrated comparable accuracy (85.29\%) and a higher recall (90.72\%) at the cost of precision (79.04\%), suggesting its potential for safety critical systems where failing to detect a yielding event is more dangerous than issuing a false alert.

While LR achieved a respectable accuracy of 88.08\%, its precision (83.66\%) and recall (90.72\%) indicate that it remains a strong statistical baseline in relatively static environments. In contrast, GPT-4o-mini recorded the lowest performance, with an accuracy of 70.20\%, recall of 37.97\%, and precision of 87.38\%. Although its high precision implies a low false positive rate, the consistently low recall across intersections limits its applicability in safety critical or high-complexity scenarios.

Site specific analysis reveals that the performance gap between models becomes more pronounced in complex environments. At Site 3, which exhibits intricate pedestrian–vehicle interactions, the shape of the intersection is T-shaped, the lack of marked crosswalks, the presence of numerous roadside parking spaces, and the concentration of nearby restaurants contribute to a high frequency of pedestrian crossings and increased environmental complexity. In this setting, GPT-4o outperformed all other models with an accuracy of 86.32\% and a perfect recall of 100\%, showcasing its robustness in handling complex dynamics. In contrast, GPT-4o-mini and LR struggled, with GPT-4o-mini achieving only 56.41\% accuracy and 48.19\% recall. Deepseek-R1, though achieving a high precision of 88.52\%, suffered from lower recall (65.06\%), indicating a conservative prediction strategy potentially driven by context sensitivity.

A similar pattern was observed at Site 16, another highly dynamic location. The shape of the intersection is four-way. This intersection is situated near a dense cluster of commercial facilities, including numerous shops and restaurants, and features roadside parking directly along the main thoroughfare. These characteristics contribute to a high volume of pedestrian activity and create a more intricate environment for pedestrian–vehicle interactions. In this setting, GPT-4o again led in performance, achieving 88.81\% accuracy and 91.92\% precision. Deepseek-V3 also performed well, with a recall of 87.88\% and a precision of 93.55\%, although its slightly elevated false positive rate suggests a trade-off inherent in aggressive detection strategies. GPT-4o-mini, however, once again exhibited very low recall (29.29\%), reaffirming its limitations in environments that demand real-time situational awareness and context-sensitive reasoning.

At simpler sites such as Site 5, models including GPT-4o, Deepseek-V3, and Deepseek-R1 achieved perfect accuracy and recall (100\%), suggesting that these LLMs generalize effectively to low complexity conditions. At Site 7, GPT-4o again achieved perfect scores, while Deepseek-R1 and Deepseek-V3 continued to show high recall with some drop in precision, indicating a more recall-favoring approach.

Meanwhile, LR performed well in certain low complexity scenarios, for example, at Site 4, it achieved an accuracy of 96.67\%, though recall was limited to 50\%. However, its performance degraded in more complex intersections, again emphasizing its limitation in environments that require nuanced reasoning. GPT-4o-mini consistently demonstrated low recall across challenging sites, reaffirming that it is more suitable for resource-constrained, low-risk applications.

In summary, GPT-4o consistently outperforms other models across a variety of traffic scenarios, offering high adaptability, accuracy, and contextual understanding. Deepseek-V3 is well suited for applications that prioritize precision and false alarm avoidance, while Deepseek-R1 is advantageous in safety critical environments where high recall is essential. LR and GPT-4o-mini, though computationally efficient, are better suited to low complexity deployments, as they lack the robustness required for dynamic, real-world applications.

\begin{figure}[tp]
	\centering
	{\includegraphics[width=.5\linewidth]{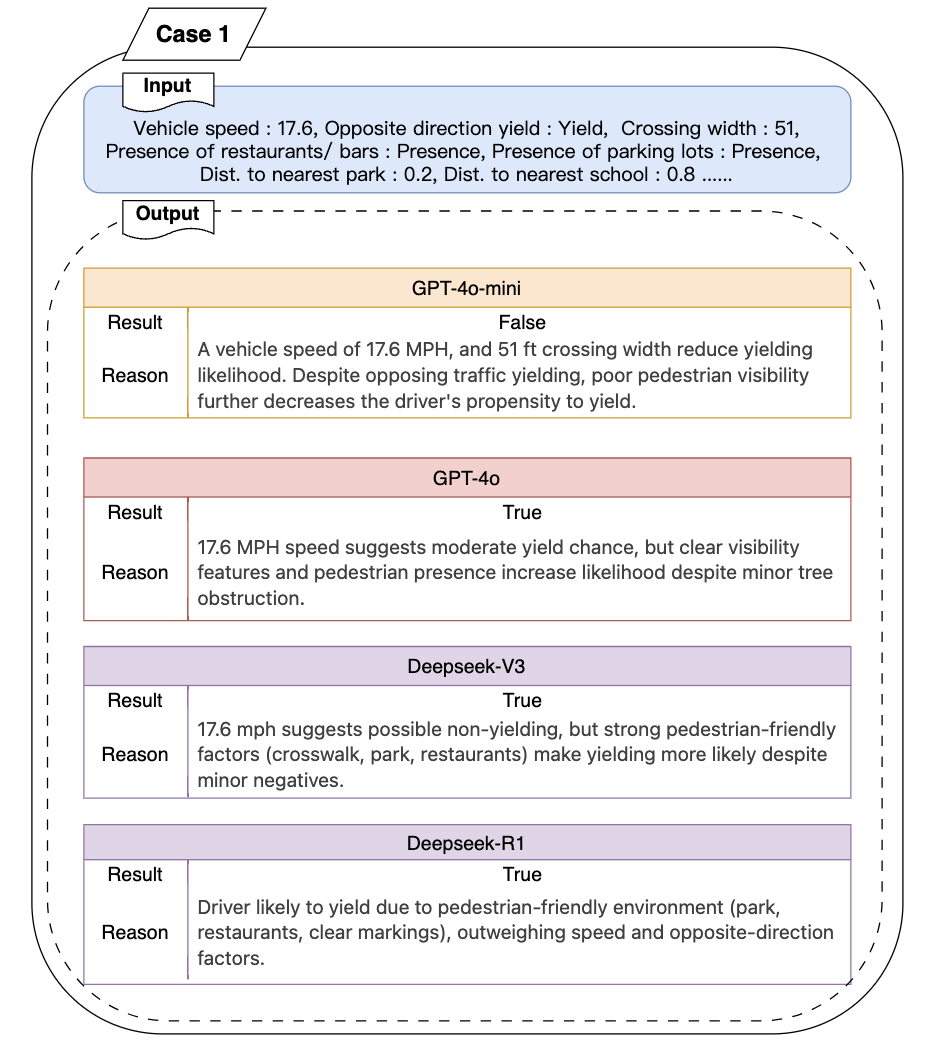}}%
	\hfil%
	{\includegraphics[width=.5\linewidth]{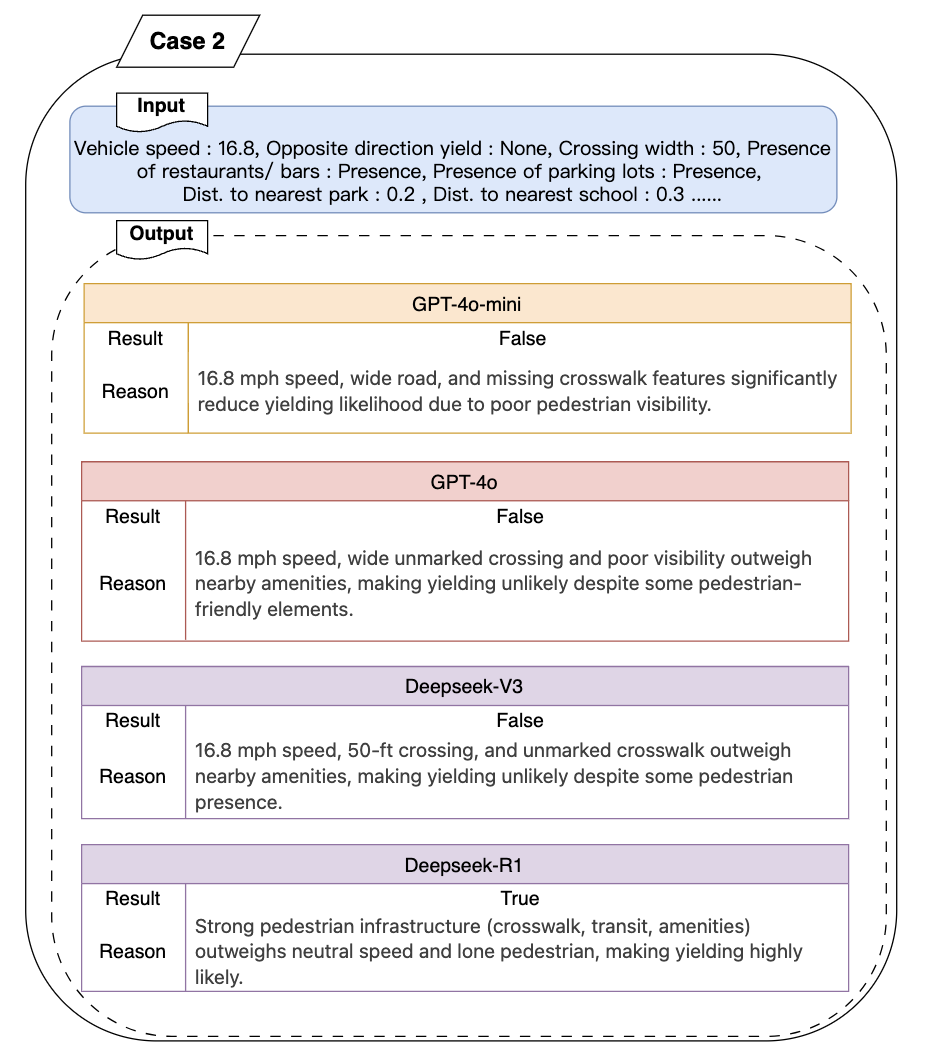}}%
	\caption{
    Two representative prediction cases given by various LLMs. The reasons to infer the LLMs’ predictions are also provided. Note that Case 1 and Case 2 have ground truths of True and False, respectively, both misclassified by LR. 
    }
	\label{Case}
\end{figure}

\subsection{Interpretability}

In addition to its superior predictive performance, GPT-4o offers significant interpretability advantages, distinguishing it from both traditional statistical models. While classical models such as LR provide feature coefficients that offer a degree of transparency, they often fall short in capturing complex feature interactions and typically lack the capacity to provide human-readable rationales for individual predictions.

By contrast, LLMs like GPT-4o are inherently more interpretable due to their ability to generate natural language explanations alongside their predictions. When provided with structured input data, GPT-4o can not only identify salient contextual features but also reason about their relative importance and produce a coherent, step-by-step justification for its decisions. This capacity for human aligned reasoning offers an intuitive and accessible interface for researchers, practitioners, and domain experts seeking to understand and validate the model’s behavior.

Fig.~\ref{Case} presents a selection of representative cases comparing model predictions across different approaches. In each of these examples, the traditional LR model failed to correctly predict whether the vehicle yielded to the pedestrian, whereas GPT-4o and most other LLMs produced accurate predictions. These instances demonstrate the advanced reasoning and generalization capabilities of LLMs, particularly in situations that require the integration of nuanced contextual information. The explanations generated by GPT-4o draw upon multiple layers of input, including vehicle speed, the presence or absence of road markings, traffic flow patterns, and surrounding environmental conditions, to emulate expert level decision-making. This process enhances both the interpretability and transparency of the model’s output and reinforces its suitability for real-world applications, where reliability and contextual understanding are critical.

\subsection{Computational cost and time analysis}

The estimated computational costs and the inference time for a single event of the adopted LLMs are presented in Table~\ref{Cost of Model}. The cost estimation is based on standard pricing per million tokens for input, cached input, and output tokens. Among the models evaluated, GPT-4o incurs the highest computational cost (\$5.31), primarily due to its greater token usage and higher pricing, which reflect its superior performance and model complexity. In contrast, GPT-4o-mini offers a more cost effective alternative (\$1.30), making it suitable for real-time applications with limited computational resources. Notably, both GPT-4o and GPT-4o-mini support multimodal input, integrating numerical, image, and textual data, which enhances their ability to extract contextual cues from complex traffic scenes. On the other hand, Deepseek-R1 and Deepseek-V3 adopt an unimodal strategy, leveraging only numerical and text inputs, yet still demonstrate strong performance in specific tasks. Among them, Deepseek-V3 shows the lowest estimated cost (\$0.21), highlighting its efficiency for large scale, high precision applications. Although Deepseek-R1 is relatively more expensive (\$2.59), it offers higher recall, making it particularly suitable for safety critical scenarios where missing a positive instance could have severe consequences. Overall, these results highlight the trade-off between predictive performance, input modality complexity, and operational cost, providing practical insights for model selection under varying deployment constraints.
Among the evaluated models, GPT-4o, GPT-4o-mini, and DeepSeek-V3 are architected as chat-oriented models, whereas DeepSeek-R1 is specifically designed as a reasoning-focused model. As a result, DeepSeek-R1 demonstrates longer inference times per event compared to the chat-based variants, making it less appropriate for real-time or low-latency application scenarios. Local deployment of open-source LLMs, which circumvents network latency inherent in API-based interactions, is expected to substantially reduce per-event inference time across all model types, including DeepSeek-R1.

\begin{table}[!ht]
    \belowrulesep=0pt
    \aboverulesep=0pt
    \renewcommand{\arraystretch}{1.5}
    \centering
    \caption{Estimated computational cost in US dollars and inference time in seconds per single prediction event across various LLMs for driver yielding behavior prediction.}
    \resizebox{\textwidth}{!}{
    \begin{tabular}{ccccccc}
    \hline
    Model& 1M input tokens& 1M cached input tokens & 1M output tokens & Est. number of tokens& Est. cost &  Inference time (s)\\
    \hline
    GPT-4o-mini& 0.15& 0.08& 0.60& 10,800,000& 1.30 & 6\\
    \hline
    GPT-4o& 2.50& 1.25& 10.00& 4,000,000& 5.31 & 7\\
    \hline
    Deepseek-V3& 0.28& 0.07& 1.12& 1,700,000& 0.21 & 9\\
    \hline
    Deepseek-R1& 0.56& 0.14& 2.23& 2,400,000& 2.59 & 210\\
    \hline
    \end{tabular}
    }
    \label{Cost of Model}
\end{table}

\section{Discussion}\label{Discussion}

Table~\ref{Summary of Model} provides a comparative summary of the five models evaluated for pedestrian safety prediction: LR, GPT-4o-mini, GPT-4o, Deepseek-V3, and Deepseek-R1. Each model exhibits distinct performance characteristics and practical trade-offs, making them suitable for different real-world scenarios.

LR serves as a simple linear baseline, offering efficient and fast inference. Its low computational cost is well suited for basic traffic safety systems in environments with limited variability, such as low traffic intersections. However, its performance degrades in complex, non-linear settings, limiting its effectiveness where nuanced interactions are common.

GPT-4o-mini, a lightweight variant of the GPT-4o model, is optimized for real-time inference with minimal resource consumption. While it delivers reasonable performance in constrained settings, its precision and recall decline in more complex scenarios. This model is most appropriate for real-time applications where rapid response and computational efficiency are paramount, and where moderate prediction accuracy is acceptable, such as in low risk zones with predictable pedestrian behavior.

GPT-4o, the full scale multimodal variant, demonstrates superior performance in both accuracy and generalization. Capable of processing both textual and visual data, it is particularly effective in complex environments requiring a nuanced understanding and multimodal reasoning. However, its high computational demands and relatively slower inference speed constrain its deployment in latency sensitive or resource limited systems. Its optimal use cases include automated driving or advanced urban safety systems that demand robust decision-making and context aware processing.

Deepseek-V3, an open source model, achieves notably high precision, effectively minimizing false positives. This reliability makes it highly suitable for safety critical applications where over-alerting could reduce trust or cause operational inefficiencies. Nevertheless, its relatively low recall means that some true pedestrian instances may be missed, which poses a limitation in applications where comprehensive detection is required.

Deepseek-R1, a reasoning oriented model, is explicitly designed to maximize recall, thereby excelling at identifying true positives even in ambiguous scenarios. Despite its slower inference and higher token level processing cost, it is particularly advantageous in high risk or complex traffic environments where missing critical pedestrian interactions could result in severe safety consequences. Its capabilities align well with applications that require deep semantic understanding and thorough situational analysis, such as intersections with a high density of pedestrian-vehicle interaction.

In summary, model selection must be guided by the specific requirements of the deployment context. LR is optimal for simple environments and a large number of training sets, GPT-4o-mini for resource-constrained scenarios, GPT-4o for high-accuracy multimodal systems, Deepseek-V3 for precision-critical deployments, and Deepseek-R1 for recall-focused applications. Understanding these trade-offs is essential for designing effective and optimized pedestrian safety systems in diverse urban settings.

\section{Limitations, challenges, and future work}\label{Future}

This paper demonstrates the potential of the proposed LLM-based approach for understanding and predicting driver–pedestrian interactions at unsignalized intersections. However, a primary limitation is the extensive preprocessing required for video data. Raw footage collected via the TIM platform must be manually annotated by trained personnel to identify and extract relevant behavioral events, which are then converted into structured textual and visual inputs. Practical, real-time deployment would require near-instantaneous inference and advanced computational hardware. These labor-intensive and time-consuming procedures make real-time deployment impractical. Facilitating the efficient transformation of real-time video data into textual narratives with per-event imagery, rather than static images per intersection, would substantially enhance the practicality and scalability of the proposed approach in real-world applications.

Moreover, challenges persist in adapting LLMs for real-time pedestrian safety applications. Reliance on API-based access introduces concerns regarding data privacy and increased latency, which may be unsuitable for latency-sensitive traffic scenarios. Real-time safety systems in urban traffic environments often demand models capable of processing raw sensor data locally and producing reliable, low-latency predictions. Achieving a high level of responsiveness and robustness constitutes a critical technical barrier to the integration of LLMs into ITS.

Therefore, our future work aims to integrate real-time data processing pipelines into LLMs by automatically transforming traffic video data into structured inputs compatible with LLMs, enabling deployment in practical traffic safety systems.
In addition, we plan to fine-tune open-source LLMs with domain-specific pedestrian safety datasets to further enhance model performance and facilitate local deployment, thereby strengthening data privacy protections and mitigating network latency by eliminating reliance on external API-based data transmission.

These advancements shall establish the foundation for early warning systems targeting drivers or pedestrians, enhancing situational awareness and safety outcomes, and thereby facilitating the development of responsive, reliable, and scalable AI systems for complex real-world traffic environments.

\begin{table}
    \aboverulesep=0pt
    \renewcommand{\arraystretch}{1.0}
    \centering
    \caption{Summary of adopted predictive models.}
    \resizebox{\textwidth}{!}{
    \begin{tabular}{ccccc}
    \hline
    Model& Description& Multimodal Prompting& Characteristics& Possible Real-World Application \\
    \hline
    LR& \begin{tabular}{p{3cm}}
    A simple linear model for binary classification.
    \end{tabular}& 
    \begin{tabular}{p{3cm}}
    Numerical inputs only.
    \end{tabular}& \begin{tabular}{p{3cm}}
    Fast, simple, low computational cost. Low performance in complex scenarios.
    \end{tabular}& \begin{tabular}{p{3cm}}
    Basic traffic safety systems with less complex scenarios, suitable for large datasets where there is sufficient data for training.
    \end{tabular} \\
    \hline
    GPT-4o-mini& \begin{tabular}{p{3cm}}
    GPT-4o lightweight version with 8B \citep{Abacha2024} parameters, closed source.
    \end{tabular}& \begin{tabular}{p{3cm}}
    Numerical, image, and text inputs.
    \end{tabular}& \begin{tabular}{p{3cm}}
    Fast inference, low resource consumption. Lower recall and accuracy in complex cases.
    \end{tabular}& \begin{tabular}{p{3cm}}
    Real-time pedestrian safety in controlled environments.
    \end{tabular} \\
    \hline
    GPT-4o& \begin{tabular}{p{3cm}}
    A multimodal chat model with 200B \citep{Abacha2024} parameters, closed source.
    \end{tabular}&\begin{tabular}{p{3cm}}
    Numerical, image, and text inputs.
    \end{tabular}& \begin{tabular}{p{3cm}}
    High accuracy, robust performance, and versatile. High computational cost, slower inference.
    \end{tabular}& \begin{tabular}{p{3cm}}
    Complex traffic environments require both recall and precision, autonomous driving systems.
    \end{tabular} \\
    \hline
    Deepseek-V3& \begin{tabular}{p{3cm}}
    A chat model, with 671B parameters, is open source.
    \end{tabular}&\begin{tabular}{p{3cm}}
    Numerical and text inputs.
    \end{tabular}& \begin{tabular}{p{3cm}}
    High precision, very reliable in minimizing false positives. Lower recall, misses some true positives.
    \end{tabular}& \begin{tabular}{p{3cm}}
    Automated safety systems with low tolerance for false positives, urban safety monitoring.
    \end{tabular} \\
    \hline
    Deepseek-R1& \begin{tabular}{p{3cm}}
    A reasoning model, with 671B parameters, is open source.
    \end{tabular}&\begin{tabular}{p{3cm}}
    Numerical and text inputs.
    \end{tabular}& \begin{tabular}{p{3cm}}
    High recall, strong at detecting true positives. Slow inference, higher token cost, lower precision.
    \end{tabular}& \begin{tabular}{p{3cm}}
    Safety-critical systems where recall is prioritized over precision, and complex environments requiring deep reasoning.
    \end{tabular} \\
    \hline
    \end{tabular}
    }
    \label{Summary of Model}
\end{table}

\section{Conclusion}\label{Conclusion}
In the evolving landscape of ML, establishing a paradigm that integrates LLMs is essential for achieving more accurate predictions and informed decision-making for safer urban mobility. This paper demonstrates a successful application of multimodal LLMs in modeling driver yielding behavior by leveraging heterogeneous traffic data and a novel prompt design. The proposed method is capable of explicating intermediate reasoning processes, enabling researchers to systematically examine how specific environmental cues and contextual factors influence yielding predictions.
We evaluate the performance of state-of-the-art LLMs from the GPT and DeepSeek series in comparison with traditional ML methods, focusing on predictive accuracy, interpretability, and computational efficiency.

The results demonstrate that LLMs, when guided by carefully designed domain-specific prompts and enriched contextual knowledge, achieve strong predictive performance in modeling driver yielding behavior. The integration of visual data and the use of step-wise reasoning chains in prompt design further enhance model interpretability and robustness, reinforcing the viability of LLM-based approaches for complex traffic safety analysis.
Specifically, GPT-4o attains the highest overall accuracy and recall, whereas DeepSeek-V3 delivers superior precision, highlighting their complementary potential for distinct safety-critical applications. 

Ultimately, this study underscores the potential of LLMs as powerful tools for advancing traffic safety analysis. It establishes a foundation for developing LLM-based, real-time intersection alert systems, with the overarching goal of improving pedestrian safety.

\clearpage

\bibliographystyle{elsarticle-harv}
\bibliography{refs}
\end{document}